\documentclass[sigconf]{acmart}

\usepackage{microtype}
\usepackage{graphicx}
\usepackage{booktabs}
\usepackage{multirow}
\usepackage{amsmath}
\usepackage{mathtools}
\usepackage{enumitem}
\usepackage{algorithm}
\usepackage{algpseudocode}
\usepackage{subcaption}
\usepackage{makecell}
\usepackage{xcolor}
\usepackage{pgfplots}
\usepackage{pgfplotstable}
\usepackage{url}
\usepackage{hyperref}

\newcommand{\PCMNO}{\textsc{PCM-NO}}

\theoremstyle{plain}
\newtheorem{proposition}{Proposition}
\theoremstyle{definition}
\newtheorem{assumption}{Assumption}
\theoremstyle{remark}
\newtheorem{remark}{Remark}

\newcommand{\nf}{\textsc{N.F.}}

\newcommand{\LogDiv}{\mathrm{Div}}
\newcommand{\LogBC}{\mathrm{BC}}

\begin{document}

\title{Learning Autoregressive Neural Operators on Solver-Defined Physical Constraint Manifolds}

\author{Nannan Zhou}
\affiliation{%
  \institution{School of Computer Science}
  \city{Sichuan University, Chengdu}
  \country{China}}
\affiliation{%
  \institution{National Key Laboratory of Fundamental Algorithms and Models for Engineering Numerical Simulation}
  \city{Chengdu}
  \country{China}}
\email{zhounannan@stu.scu.edu.cn}

\author{Guanmeng Xian}
\affiliation{%
  \institution{School of Computer Science}
  \city{Sichuan University, Chengdu}
  \country{China}}
\affiliation{%
  \institution{National Key Laboratory of Fundamental Algorithms and Models for Engineering Numerical Simulation}
  \city{Chengdu}
  \country{China}}
\email{xianguanmeng@stu.scu.edu.cn}

\author{Ning Yang}
\authornote{Corresponding author}
\affiliation{%
  \institution{School of Computer Science}
  \city{Sichuan University, Chengdu}
  \country{China}}
\affiliation{%
  \institution{National Key Laboratory of Fundamental Algorithms and Models for Engineering Numerical Simulation}
  \city{Chengdu}
  \country{China}}
\email{yangning@scu.edu.cn}

%\correspondingauthor
%\email{webmaster@marysville-ohio.com}
%\affiliation{%
%  \institution{Institute for Clarity in Documentation}
%  \country{USA}
%}
%
%\author{Lars Th{\o}rv{\"a}ld}
%\affiliation{%
%  \institution{The Th{\o}rv{\"a}ld Group}
%\email{larst@affiliation.org}
%\affiliation{%
%  \institution{Inria Paris-Rocquencourt}

\begin{abstract}
Autoregressive neural operators approximate time-dependent PDE solvers by
repeatedly composing a learned transition, although the resulting states need
not remain feasible under the discrete constraints of the numerical problem.
We study this issue for incompressible flow, where each velocity state must
satisfy solver-level incompressibility together with prescribed boundary
values. We formulate autoregressive operator learning as a map between
solver-defined affine feasible sets and introduce \PCMNO{}, a projected-residual
realization of this feasible-transition principle. The neural backbone predicts
a candidate state increment, a fixed tangent projection extracts its admissible
component, and a metric retraction maps the advanced state to the prescribed
feasible set. With exact affine constraint solves, the resulting transition is
closed on the feasible set, while an affine identity separates admissible-update
parameterization from state-level closure. With approximate solves,
tangent-only residuals admit a cumulative upper bound, whereas per-step
retraction resets feasibility to the residual achieved by the current solve.
Experiments on periodic Navier--Stokes flow and lid-driven cavity flow
demonstrate the superiority of \PCMNO{} in preserving solver-level constraints
over long autoregressive rollouts and under parameter, boundary-value, and
resolution shifts, while achieving lower rollout errors and better
flow-specific diagnostics across both in-distribution and stress settings.
\end{abstract}



\maketitle

\section{Introduction}
\label{sec:introduction}

Neural operators provide reusable function-to-function surrogates for
parametric PDE families and have become an effective approach to accelerating
repeated scientific simulation
\citep{lu2021learning,li2021fourier,kovachki2023neural}.
For time-dependent systems, an autoregressive neural operator learns a
short-time transition and obtains long-horizon predictions by repeatedly
composing that transition
\citep{brandstetter2022message,mccabe2023towards,lippe2023pde}.
This formulation mirrors the stepwise structure of numerical solvers and
supports flexible prediction horizons across varying physical conditions.
However, it also makes every predicted state part of the input encountered by
subsequent transitions. Constraint violations that survive one step can
therefore enter the remainder of the rollout and progressively affect its
physical validity and predictive stability.

This issue is especially acute in incompressible flow, where preservation of
local fluid volume requires a divergence-free velocity field, and a fixed
discretization encodes this requirement together with prescribed boundary
values as algebraic constraints on admissible velocity states
\citep{guermond2006overview}. Yet solver-generated training trajectories alone
do not guarantee that a learned transition preserves this admissible set from
one step to the next
\citep{li2022physics,mccabe2023towards}. Existing physics-aware approaches address constraint violations through
{soft-constraint objectives} based on residual or conservation penalties
\citep{raissi2019physics,li2022physics,karniadakis2021physics},
{structure-specific representations}
\citep{richterpowell2022neural,espath2021statistical,cheng2026decoupled},
or {post-hoc state projection}
\citep{li2026project}. 
These approaches have distinct limitations in
autoregressive rollout. \textbf{Soft-constraint objectives} do not guarantee
feasibility at every step, so residual violations can enter subsequent
predictions and accumulate over time. \textbf{Structure-specific representations} encode selected constraints but
lack composability, requiring specialized designs for joint conditions such as
incompressibility and nonhomogeneous boundary values. \textbf{Post-hoc state projection} like FinalProj \citep{li2026project} restores output feasibility only after an
unconstrained next-state prediction, without constraining the learned update
before state advancement.

These limitations motivate the feasible-transition view of autoregressive
operator learning. Rather than allowing the learned dynamics to evolve freely
in the ambient state space, \emph{we treat the solver-defined feasible set as
the state space of the learned transition, so that every autoregressive
prediction remains physically feasible at the solver level}. This set captures
discrete kinematic requirements such as incompressibility and prescribed
boundary values, while the neural operator learns the dynamics within the
feasible set.
Based on this idea, we propose
\textbf{Physical-Constraint-Manifold Neural Operators} (\textbf{\PCMNO{}}),
a unified transition-level formulation in which the solver-defined feasible
set itself serves as the state space of the learned dynamics. \PCMNO{}
realizes this formulation by coupling a backbone neural operator with two
solver-defined components. The \textbf{tangent projection} filters the
backbone's candidate residual update before state advancement, so that only
directions compatible with the homogeneous constraints enter the rollout. The
\textbf{metric retraction} then maps the advanced state to the feasible set
associated with the current constraint target, resetting numerical constraint
residuals at each step. The \textbf{advantages} of \PCMNO{} are threefold. First, its
projection--retraction factorization makes per-step solver-level feasibility a
structural property of the transition, independent of loss balancing and of
whether the backbone learns the constraints from data. Second, the
projected-residual realization exposes the admissible component entering the
learned dynamics and removes constraint-incompatible directions before state
advancement, providing an interpretable inductive bias beyond state-level
correction alone. Third, \PCMNO{} provides a unified transition formulation whose parameter-free,
solver-derived layers accommodate periodic incompressibility, joint
incompressibility--boundary constraints, and step-dependent affine targets.
%Per-step retraction further resets constraint residuals to the current solve accuracy.

Our analysis separates exact feasible-state closure, admissible-update
parameterization, and finite-solve behavior. Under exact affine constraints,
the direct-state and projected-residual realizations belong to the same
feasible-transition family and map feasible inputs to the prescribed affine
set. The tangent layer explicitly exposes the admissible component entering
the learned dynamics and removes its orthogonal constraint-incompatible
component before state advancement. Under approximate solves, tangent-only
residuals can accumulate across rollout steps, whereas retraction resets the
state residual to the accuracy achieved by the current solve and accommodates
changes in the affine target. These results characterize the geometry and
numerical behavior of the \PCMNO{} transition. The experiments then evaluate
its predictive accuracy, rollout stability, and generalization under stress.

Our contributions can be summarized as follows:
\begin{itemize}[leftmargin=*,itemsep=1pt,topsep=2pt]
    \item We formulate autoregressive operator learning as a feasible-state
    transition and propose \PCMNO{}, which couples a neural candidate-update
    backbone with successive tangent projection and metric retraction to keep
    every rollout step feasible under solver-defined affine constraints.

    \item We establish exact feasible-state closure, characterize the affine
    relationship between tangent projection and retraction, and analyze
    orthogonal update filtering together with residual accumulation and reset
    under approximate constraint solves.
    
    \item Extensive experiments on periodic and boundary-constrained
incompressible flows verify the superiority of \PCMNO{}.
%
%\item Extensive experiments on periodic and boundary-constrained
%incompressible flows show that \PCMNO{} preserves solver-level feasibility over
%long autoregressive rollouts and under parameter, boundary-value, and resolution
%shifts, while improving rollout accuracy and flow-specific diagnostics over
%soft-constraint training and post-hoc state projection. Component-level
%diagnostics further verify tangent filtering and per-step residual reset.
\end{itemize}

\section{Preliminaries}
\label{sec:prelim}

\subsection{Time-Evolution Operator and Discrete Realization}

At each time \(t\), the state
\(u_t:=u(\cdot,t)\in\mathcal X\) is a spatial field function over the physical
domain. The time-\(\Delta t\) evolution operator \(\mathcal S_{\Delta t}\) maps the field function \(u_t\) to the
one at the next time level:
\begin{equation}
\label{eq:continuous_transition}
\mathcal S_{\Delta t}(u_t;\mu,g)
=
u_{t+\Delta t},
\end{equation}
where \(\mu\) denotes physical parameters and \(g\) boundary or forcing data.
%It is therefore an operator rather than a finite-dimensional pointwise map,
%because both its input and output are spatial functions. 
At spatial resolution
\(h\), these fields are represented by discrete states
\(u_{h,k}\in\mathcal X_h\) at \(t_k=t_0+k\Delta t\), and the reference solver
induces the discrete evolution operator
\begin{equation}
\label{eq:discrete_transition}
u_{h,k+1}
=
\mathcal S_{h,\Delta t}(u_{h,k};\mu,g_h).
\end{equation}
The learned transition is defined on this same discrete state representation,
where the solver-level constraints are specified.

\subsection{Feasible States}
\label{sec:discrete_feasible_states}

Let \(m_h\) be the number of scalar constraints. At a fixed discretization and
conditioning, a discrete state \(u_h\in\mathcal X_h\) is called
\emph{feasible} if it satisfies the solver-defined constraint system
\(C_hu_h=b_h\), where \(C_h:\mathcal X_h\to\mathbb R^{m_h}\) is the discrete constraint
operator, and \(b_h\in\mathbb R^{m_h}\) contains the corresponding prescribed
values.  The set of all feasible states is
\begin{equation}
    \mathcal M_h(b_h)
    =
    \left\{
        u_h\in\mathcal X_h:
        C_hu_h=b_h
    \right\}.
    \label{eq:feasible_set}
\end{equation}
The operator \(C_h\) may combine multiple solver-defined constraints.
For incompressible flow, it includes the discrete divergence operator \(D_h\)
as the block encoding incompressibility, while the corresponding block of
\(b_h\) is zero. Consequently, every feasible state
\(u_h\in\mathcal M_h(b_h)\) satisfies \(D_hu_h=0\), the discrete counterpart
of a divergence-free velocity field. Both \(C_h\) and \(b_h\) are determined
by the numerical problem and are not learnable.

We assume that \(C_h\) is linear and that
\(\mathcal M_h(b_h)\) is nonempty for every problem instance considered.
Under these conditions, the feasible set is an affine translate of
\(\ker C_h\) and therefore an embedded affine manifold. We call the algebraic
object in Eq.~\eqref{eq:feasible_set} the
\emph{solver-defined affine feasible set} and its geometric interpretation the
\emph{physical constraint manifold}.

\begin{figure*}[t]
    \centering
    \includegraphics[
        width=0.75\textwidth,
        height=0.28\textheight,
%        keepaspectratio
    ]{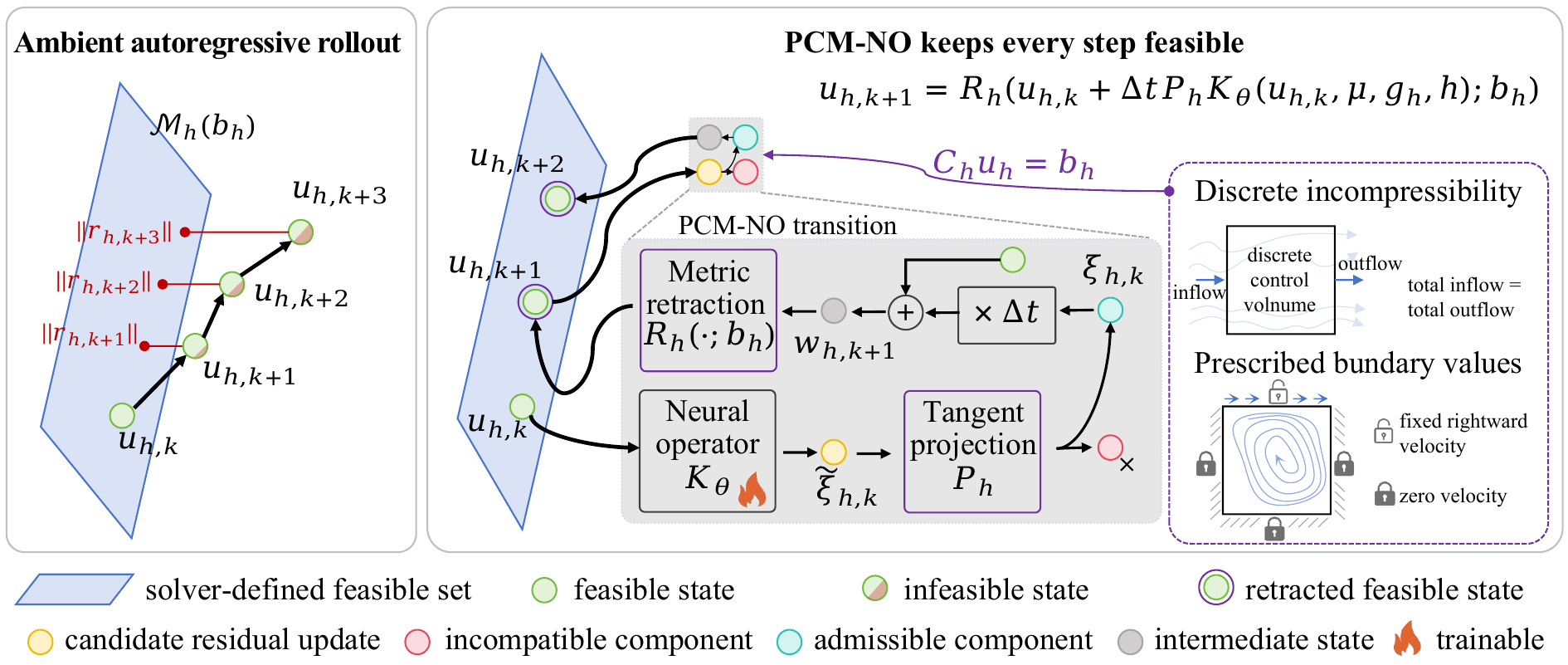}
    \caption{
    Overview of \PCMNO{}. An ambient autoregressive rollout can propagate
    constraint-incompatible components across steps. In \PCMNO{}, the backbone
    predicts a candidate update, \(P_h\) retains its admissible component, and
    \(R_h\) returns the advanced state to the prescribed affine feasible set.
    The right panel illustrates solver-defined constraints encoded by
    \(C_hu_h=b_h\).
    }
    \label{fig:pcmno_overview}
\end{figure*}

%\begin{figure*}[t] 
%\centering 
%\includegraphics[width=0.662\textwidth] {figures/pcmno_overview.png} 
%\caption{ Overview of \PCMNO{}. The backbone predicts a candidate update direction, \(P_h\) removes its constraint-incompatible component, and \(R_h\) retracts the advanced state to the prescribed affine feasible set. The right panel shows the P-NS and LC constraint configurations. } 
%\label{fig:pcmno_overview} 
%\end{figure*}

\subsection{Problem formulation}
\label{sec:problem_formulation}
Let the \(i\)-th trajectory
$
    U_h^{(i)}
    =
    \bigl(u^{(i)}_{h,0},u^{(i)}_{h,1},\ldots,u^{(i)}_{h,T_i}\bigr)
$
be generated by the reference solver under physical parameters
\(\mu^{(i)}\), discretized boundary or forcing data \(g_h^{(i)}\), and the
corresponding constraint target \(b_h^{(i)}\). 
%Here, \(g_h^{(i)}\) denotes exogenous problem data represented on the spatial discretization and may determine part of \(b_h^{(i)}\). 
The training set contains adjacent conditioned function pairs
\[
    \left\{
    \bigl((u^{(i)}_{h,k},\mu^{(i)},g_h^{(i)}),\,u^{(i)}_{h,k+1}\bigr)
    :
    i=1,\ldots,n,\;
    k=0,\ldots,T_i-1
    \right\},
\]
with \(u^{(i)}_{h,k},u^{(i)}_{h,k+1}\in\mathcal M_h(b_h^{(i)})\), where \(n\) is the number of training trajectories and \(T_i\) is the number of transitions in the \(i\)-th trajectory.

The goal is to learn a neural operator \(\Phi_\theta\) that approximates the solver-induced transition while preserving the feasible state set:
\begin{equation}
\label{eq:problem_goal}
\begin{aligned}
\Phi_\theta(\cdot;\mu,g_h,h)
&:
\mathcal M_h(b_h)
\rightarrow
\mathcal M_h(b_h),
\\
\Phi_\theta(u_{h,k};\mu,g_h,h)
&\approx
\mathcal S_{h,\Delta t}(u_{h,k};\mu,g_h).
\end{aligned}
\end{equation}
The first relation in Eq.~\eqref{eq:problem_goal} specifies a structural
closure requirement for the learned transition, while the second relation defines
the statistical approximation objective.

\section{Method}
\label{sec:method}

\subsection{Overview}
\label{sec:method_overview}

Figure~\ref{fig:pcmno_overview} contrasts an ambient autoregressive rollout
with the feasible-state transition of \PCMNO{} and shows how its fixed layers
are derived from solver-defined constraints. In the ambient rollout, each
predicted state becomes the input to the next transition, so any
constraint-incompatible component can be propagated and amplified across
steps. \PCMNO{} instead factorizes each learned residual transition into a
candidate update, tangent projection, state advancement, and metric
retraction. This construction separates data-driven dynamics from
solver-level constraint enforcement, allowing the backbone to learn within the
feasible state space defined by \(C_hu_h=b_h\). The same formulation therefore
supports both periodic incompressibility and joint
incompressibility--boundary constraints.

Given a feasible state \(u_{h,k}\in\mathcal M_h(b_h)\), the backbone predicts a
candidate update in the ambient discrete space. The tangent projection \(P_h\)
retains its admissible component in \(\ker C_h\) before state advancement, and
the metric retraction \(R_h\) returns the advanced state to the affine feasible
set associated with the current constraint target:
\begin{equation}
\label{eq:pcmno_transition}
\begin{aligned}
\Phi_\theta(u_{h,k};\mu,g_h,h)
&=
\Bigl[
R_h(\cdot;b_h)
\circ
\bigl(
\operatorname{Id}
+
\Delta t\,P_hK_\theta(\cdot,\mu,g_h,h)
\bigr)
\Bigr](u_{h,k})
\\
&=
R_h\!\left(
u_{h,k}
+
\Delta t\,P_hK_\theta(u_{h,k},\mu,g_h,h);
b_h
\right),
\end{aligned}
\end{equation}
where \(\operatorname{Id}\) denotes the identity operator on the
discrete state space \(\mathcal X_h\).
Note that only \(K_\theta\) is trainable. The projection \(P_h\) filters the learned
update, while the retraction \(R_h\) resets state-level feasibility at every rollout
step. Both layers are constructed from the discrete constraint system
\(C_hu_h=b_h\), allowing the same transition-level formulation to accommodate
constraints such as discrete incompressibility and prescribed boundary values.

\subsection{Tangent projection and retraction}
\label{sec:tangent_projection}

Let \(W_h\succ0\) define the discrete inner product
$\langle a_h,b_h\rangle_{W_h}=a_h^\top W_h b_h$, $ \|a_h\|_{W_h}^2=a_h^\top W_h a_h$.
For \(u_h\in\mathcal M_h(b_h)\), let 
\(\mathcal U_h := T_{u_h}\mathcal M_h(b_h)\) denote the tangent space of
\(\mathcal M_h(b_h)\) at \(u_h\), namely the set of admissible
directions along which the constraint value remains unchanged.
Because \(C_h\) is linear,
\[
C_h(u_h+\tau\xi_h)
=
b_h+\tau C_h\xi_h.
\]
Hence \(u_h+\tau\xi_h\) remains in \(\mathcal M_h(b_h)\) precisely
when \(C_h\xi_h=0\). The tangent space is therefore independent of
the base point and satisfies
\begin{equation}
\label{eq:tangent_space}
    \mathcal U_h
    =
    \left\{
        \xi_h\in\mathcal X_h:
        C_h\xi_h=0
    \right\}
    =
    \ker C_h,
    \quad
    u_h\in\mathcal M_h(b_h).
\end{equation}

\paragraph{Tangent projection.}
Given a candidate update \(v_h\in\mathcal X_h\), the tangent projection is the \(W_h\)-orthogonal projection onto \(\mathcal U_h\):
\begin{equation}
\label{eq:projection_def}
    P_h v_h
    =
    \operatorname*{arg\,min}_{\xi_h\in\mathcal U_h}
    \frac{1}{2}\|\xi_h-v_h\|_{W_h}^{2}.
\end{equation}
Thus the projected update \(P_hv_h\in\ker C_h\), which can be added to a feasible state without changing the constraint value.

\paragraph{Metric retraction.}
For an intermediate state \(w_h\in\mathcal X_h\), we define the metric
retraction \(R_h\) as the \(W_h\)-metric projection onto the affine feasible
set:
\begin{equation}
\label{eq:state_projection}
    R_h(w_h;b_h)
    =
    \operatorname*{arg\,min}_{u_h\in\mathcal M_h(b_h)}
    \frac{1}{2}
    \|u_h-w_h\|_{W_h}^{2}.
\end{equation}
%Equivalently, using the definition of \(\mathcal M_h(b_h)\),
%\[
%    R_h(w_h;b_h)
%    =
%    \operatorname*{arg\,min}_{u_h\in\mathcal X_h}
%    \frac{1}{2}
%    \|u_h-w_h\|_{W_h}^{2}
%    \quad
%    \mathrm{s.t.}
%    \quad
%    C_hu_h=b_h.
%\]
For a feasible base state \(u_h\in\mathcal M_h(b_h)\) and a tangent
increment \(\delta u_h\in\mathcal U_h\), the corresponding local map
\[
    \mathcal R_{h,u_h}(\delta u_h)
    :=
    R_h(u_h+\delta u_h;b_h)
\]
satisfies
$  \mathcal R_{h,u_h}(0)=u_h$ and $\mathcal R_{h,u_h}(\delta u_h)=u_h+\delta u_h$.

\subsection{\PCMNO{} Rollout Transition}
\label{sec:pcmno_operator}

We now specify the projected residual transition summarized in
Figure~\ref{fig:pcmno_overview}. At time level \(k\), the backbone produces a
candidate update
\[
\widetilde\xi_{h,k}
=
K_\theta(u_{h,k},\mu,g_h,h),
\qquad
\widetilde\xi_{h,k}\in\mathcal X_h .
\]
The tangent layer filters this update and advances the current state according
to
$
\xi_{h,k}
=
P_h\widetilde\xi_{h,k}$ and
$
w_{h,k+1}
=
u_{h,k}
+
\Delta t\,\xi_{h,k}$.
The complete \PCMNO{} transition is
\begin{equation}
\label{eq:pcmno_update}
\Phi_\theta(u_{h,k};\mu,g_h,h)
=
R_h\!\left(
u_{h,k}
+
\Delta t\,
P_hK_\theta(u_{h,k},\mu,g_h,h);
b_h
\right).
\end{equation}

The backbone determines a candidate update
\(\widetilde{\xi}_{h,k}\), while only
\(\xi_{h,k}=P_h\widetilde{\xi}_{h,k}\) enters the rollout. If
\(u_{h,k}\in\mathcal M_h(b_h)\) and the tangent projection \(P_h\) is
exact, then the intermediate state is already feasible, i.e., 
\(w_{h,k+1}\in\mathcal M_h(b_h)\). In this case, the final retraction acts
as the identity,
\(R_h(w_{h,k+1};b_h)=w_{h,k+1}\), leaving a feasible
intermediate state unchanged. When the tangent solve is approximate or the affine target changes between
steps, \(R_h\) becomes active by resetting state-level feasibility and mapping
the state to the prescribed next feasible set.

\paragraph{Relation to post-hoc state projection.}
The feasible-transition formulation subsumes post-hoc state projection
\citep{li2026project} as a direct-state realization. Let
\(G_\theta:\mathcal X_h\to\mathcal X_h\) be an unconstrained next-state
predictor and define
$
    K_\theta^{G}(u_h)
    =
    \frac{G_\theta(u_h)-u_h}{\Delta t}.
$
For \(u_h\in\mathcal M_h(b_h)\), the exact affine projection identity gives
\begin{align}
    R_h\!\left(G_\theta(u_h);b_h'\right)
    &=
    R_h\!\left(
        u_h+\Delta t\,K_\theta^{G}(u_h);
        b_h'
    \right)
    \nonumber\\
    &=
    R_h\!\left(
        u_h+\Delta t\,P_hK_\theta^{G}(u_h);
        b_h'
    \right).
    \label{eq:posthoc_instance}
\end{align}
Thus, post-hoc state projection is the direct-state realization of the affine
feasible-transition principle, while \PCMNO{} is its projected-residual
realization. Both inherit exact closure, while \PCMNO{} makes the admissible
update explicit.

\paragraph{Changing affine targets.}
When the constraint operator \(C_h\) remains fixed but its right-hand side
changes from \(b_{h,k}\) to \(b_{h,k+1}\), the same tangent projection can be
combined with a retraction onto the next feasible set:
\begin{equation}
\label{eq:pcmno_changing_target}
\Phi_\theta(u_{h,k};\mu,g_{h,k},h)
=
R_h\!\left(
u_{h,k}
+
\Delta t\,
P_hK_\theta(u_{h,k},\mu,g_{h,k},h);
b_{h,k+1}
\right).
\end{equation}
Thus, a state in \(\mathcal M_h(b_{h,k})\) is mapped to a state in
\(\mathcal M_h(b_{h,k+1})\), while the homogeneous admissible direction space
remains \(\ker C_h\).

\subsection{Training objective}

Training minimizes prediction error over autoregressive rollout windows. Let
\(H\) denote the training horizon and \(s\) the starting index of a window.
For trajectory \(i\), initialize
$
    \widehat u^{(i,s)}_{h,0}=u^{(i)}_{h,s},
$
and recursively apply
\[
    \widehat u^{(i,s)}_{h,r+1}
    =
    \Phi_\theta\!\left(
        \widehat u^{(i,s)}_{h,r};
        \mu^{(i)},g_h^{(i)},h
    \right),
    \qquad r=0,\ldots,H-1 .
\]
The reported models are trained with
\begin{equation}
\label{eq:rollout_loss}
    \mathcal L_{\mathrm{train}}(\theta)
    =
    \sum_{i=1}^{n}
    \sum_{s=0}^{T_i-H}
    \sum_{r=1}^{H}
    \omega_r\,
    \ell_h\!\left(
        \widehat u^{(i,s)}_{h,r},
        u^{(i)}_{h,s+r}
    \right),
\end{equation}
where \(\ell_h\) is a discrete field-level prediction loss and \(\omega_r\)
controls the relative contribution of rollout step \(r\). Note that the feasibility is specified by the transition
architecture in Eq.~\eqref{eq:pcmno_update}. 

\section{Structural Analysis}
\label{sec:theory}

The analysis characterizes the solver-defined tangent projection and metric
retraction, as well as the residual behavior of their numerical realizations. All proofs and derivations are given in Appendix~\ref{app:proofs}.

%\begin{assumption}[Fixed affine constraint model]
%\label{ass:fixed_affine}
%The state space \(\mathcal X_h\) is finite-dimensional,
%\(C_h:\mathcal X_h\to\mathbb R^{m_h}\) is linear,
%\(\mathcal M_h(b_h)\) is nonempty, and \(W_h\succ0\) defines the
%discrete metric used by the tangent and state projections. Unless stated
%otherwise, both constrained subproblems are solved exactly.
%\end{assumption}

\begin{proposition}[Exact feasible-state transition]
\label{prop:exact_closure}
Any feasible input
\(u_{h,k}\in\mathcal M_h(b_h)\) satisfies
\begin{equation}
    \Phi_\theta(u_{h,k};\mu,g_h,h)
    \in
    \mathcal M_h(b_h).
\end{equation}
If \(\widehat u_{h,0}\in\mathcal M_h(b_h)\), then the exact
autoregressive rollout remains in \(\mathcal M_h(b_h)\) for every
\(k\ge0\).
More generally, when the target changes between steps,
\[
\begin{aligned}
u_{h,k}\in\mathcal M_h(b_{h,k})
\quad\Longrightarrow\quad
&
R_h\!\left(
    u_{h,k}
    +\Delta t\,P_hK_\theta(u_{h,k});
    b_{h,k+1}
\right)
\\[-1mm]
&
\hspace{18mm}\in\mathcal M_h(b_{h,k+1}).
\end{aligned}
\]
\end{proposition}

\begin{remark}[Interpretation of exact closure]
\label{rem:closure_interpretation}
Proposition~\ref{prop:exact_closure} shows that the exact
\PCMNO{} transition maps a feasible state to the prescribed affine feasible
set, including under step-dependent targets. This closure is determined by
the fixed solver-defined constraint layers and is independent of the learned
backbone and its parameters. Sections~\ref{sec:id_feasible_rollouts}--%
\ref{sec:stress_settings} will evaluate the corresponding numerical feasibility
under in-distribution and shifted conditions.
\end{remark}

\begin{proposition}[Tangent filtering toward admissible increments]
\label{prop:tangent_filtering}
For any candidate increment
\(
\widetilde{\delta u}_h\in\mathcal X_h
\),
with
\(
\widetilde{\delta u}_h
=
\Delta t\,K_\theta(\cdot)
\)
in the \PCMNO{} transition, and any admissible target increment
\(
\delta u_h^\star\in\mathcal U_h
\),
\begin{equation}
\begin{aligned}
    \|
        \widetilde{\delta u}_h-\delta u_h^\star
    \|_{W_h}^{2}
    &=
    \|
        P_h\widetilde{\delta u}_h-\delta u_h^\star
    \|_{W_h}^{2}
    \\
    &\quad+
    \|
        (I-P_h)\widetilde{\delta u}_h
    \|_{W_h}^{2}.
\end{aligned}
\label{eq:pythagorean_filtering}
\end{equation}
Consequently,
\begin{equation}
    \|
        P_h\widetilde{\delta u}_h-\delta u_h^\star
    \|_{W_h}
    \le
    \|
        \widetilde{\delta u}_h-\delta u_h^\star
    \|_{W_h}.
\end{equation}
The target may be the difference between two states in the same affine
feasible set or the tangent projection of a reference increment.
\end{proposition}

\begin{remark}[Interpretation of tangent filtering]
\label{rem:tangent_filtering_interpretation}
Proposition~\ref{prop:tangent_filtering} shows that \(P_h\) removes exactly
the \(W_h\)-orthogonal component outside \(\mathcal U_h\) and cannot increase
the distance to any admissible target increment. When two consecutive
reference states satisfy the same affine target,
\(C_h(u_{h,k+1}-u_{h,k})=0\), so their increment lies in
\(\mathcal U_h\) up to the data-generation tolerance. Tangent filtering
therefore preserves dynamics within the feasible set while removing
constraint-incompatible components. Section~\ref{sec:mechanism_diagnostics}
verifies this decomposition on realized model updates.
\end{remark}

\begin{assumption}[Approximate numerical constraint layers]
\label{ass:approximate_layers}
At rollout step \(k\), the implemented tangent projection
\(\widehat P_{h,k}\) and metric retraction \(\widehat R_{h,k}\) satisfy
\[
    \left\|
        C_h\widehat P_{h,k}v_{h,k}
    \right\|
    \le
    \epsilon^P_{h,k},
    \quad
    \left\|
        C_h\widehat R_{h,k}(w_h;b_{h,k+1})
        -
        b_{h,k+1}
    \right\|
    \le
    \epsilon^R_{h,k}.
\]
Here, \(\epsilon^P_{h,k}\ge 0\) and
\(\epsilon^R_{h,k}\ge 0\) denote the per-step residual bounds achieved by
the numerical tangent-projection and metric-retraction solves,
respectively, which may depend on the discretization, rollout step, and numerical
solve budget.
\end{assumption}

\begin{proposition}[Residual accumulation bound and per-step reset]
\label{prop:approximate_residuals}
Let
\(
r_{h,k}=C_h\widehat u_{h,k}-b_h
\)
be the residual for a fixed target \(b_h\). A tangent-only rollout
$
    \widehat u^{\mathrm{tan}}_{h,k+1}
    =
    \widehat u^{\mathrm{tan}}_{h,k}
    +
    \Delta t\,\widehat P_{h,k}v_{h,k}
$
satisfies
\begin{equation}
    \|
        r^{\mathrm{tan}}_{h,k}
    \|
    \le
    \|
        r^{\mathrm{tan}}_{h,0}
    \|
    +
    \Delta t
    \sum_{j=0}^{k-1}
    \epsilon^P_{h,j}.
    \label{eq:tangent_residual_bound}
\end{equation}
The full transition with a final approximate metric retraction satisfies
\begin{equation}
    \|
        C_h\widehat u^{\mathrm{full}}_{h,k+1}
        -
        b_{h,k+1}
    \|
    \le
    \epsilon^R_{h,k}.
    \label{eq:retraction_reset_bound}
\end{equation}
\end{proposition}

\begin{remark}[Interpretation of residual propagation]
\label{rem:residual_propagation}
Proposition~\ref{prop:approximate_residuals} shows that residuals from
approximate tangent projections can accumulate across rollout steps, whereas
the metric retraction resets the state residual to the accuracy achieved by
the current numerical solve. This behavior will be examined empirically in
Section~\ref{sec:mechanism_diagnostics}.
\end{remark}

\section{Experiments}
\label{sec:experiments}

\subsection{Experimental setup}
\label{sec:experimental_setup}

\paragraph{Datasets and evaluation settings.}
We use two datasets governed by the 2-dimensional
incompressible Navier--Stokes equations. \textbf{Periodic NS} (\textbf{P-NS})
is forced periodic Kolmogorov flow on \([0,2\pi]^2\), with \(C_h=D_h\) and
\(b_h=0\), where \(D_h\) is the periodic spectral divergence operator.
\textbf{Lid-driven cavity flow} (\textbf{LC}) is defined on \([0,1]^2\) with
no-slip walls and a prescribed moving lid, using \(C_h=[D_h;B_h]\) and
\(b_h=[0;g_h]\), where \(B_h\) extracts boundary degrees of freedom and \(g_h\)
stores the wall and lid velocities. On these datasets, we evaluate
\textbf{in-distribution rollout}, \textbf{long-horizon rollout}, and
\textbf{stress} settings that vary Reynolds number, lid velocity, resolution,
data budget, or projection tolerance without changing the underlying dataset
definitions. Appendix~\ref{app:constraint_configurations_stress_protocol}
describes the two constraint configurations and the construction of the stress
conditions, while Table~\ref{tab:dataset_statistics} in Appendix summarizes the data
generation, discretization, and split statistics.
%\paragraph{Baselines.}
%All main methods use the same FNO backbone. \textbf{FNO} is the unconstrained
%autoregressive reference \citep{li2021fourier}. \textbf{DivReg} denotes the soft-constraint baseline,
%following the penalty-based treatment of incompressibility in physics-informed
%operator learning \citep{li2022physics}: it uses a discrete divergence penalty
%on P-NS and divergence together with prescribed-boundary penalties on LC.
%\textbf{PINO} augments the rollout objective with a dataset-specific discrete
%PDE residual in the operator-learning setting, without instance-wise test-time
%optimization \citep{li2022physics}. \textbf{FinalProj} predicts an unconstrained
%next state and applies the same feasible-set retraction before that state enters
%the next autoregressive step. On P-NS, FinalProj instantiates the deterministic
%output-projection paradigm of divergence-free neural operators
%\citep{li2026project}: it predicts an unconstrained next state and applies the
%same spectral Leray projection before the state enters the next autoregressive
%step. Our LC implementation extends this control to the joint affine constraint
%\(C_h=[D_h;B_h]\). \textbf{\PCMNO{}} predicts a candidate residual update,
%projects the update into the homogeneous constraint space, advances the state,
%and applies a final metric retraction.

\paragraph{Baselines.}
All methods use FNO-based backbones with matched capacity. \textbf{FNO} is the
unconstrained autoregressive reference \citep{li2021fourier}. \textbf{DivReg}
follows the penalty-based treatment of incompressibility in physics-informed
operator learning \citep{li2022physics}, adding a divergence penalty on P-NS
and both divergence and prescribed-boundary penalties on LC. \textbf{PINO}
augments the rollout objective with a dataset-specific discrete PDE residual,
without instance-wise test-time optimization \citep{li2022physics}. \textbf{ClawNO} is an external structure-preserving baseline evaluated only on
P-NS using its FNO-based clawFNO realization
\citep{liu2024clawno}, since its periodic potential representation does not directly
encode the stacked LC constraint \(C_h=[D_h;B_h]\). Its implementation details are
provided in Appendix~\ref{app:clawno_external_baseline}. \textbf{FinalProj} is a representative post-hoc state-projection baseline,
following the direct-state correction principle \citep{li2026project}.  On P-NS, it uses a spectral Leray projection; on LC, it
enforces the joint affine constraint \(C_h=[D_h;B_h]\).
\textbf{\PCMNO{}} instead predicts a candidate residual update, projects it
onto \(\ker C_h\), advances the state, and applies a final metric retraction.

\paragraph{Training and reporting.}
Unless stated otherwise, all methods use identical splits and autoregressive
training windows of length \(H=4\) on \(64^2\) data with
\(Re\in[100,500]\). LC additionally uses prescribed lid velocities in
\(U_{\mathrm{lid}}\in[0.8,1.2]\). Method-specific loss weights are fixed before
test evaluation. The P-NS PINO residual weight is selected using validation
rollouts, while the remaining soft-loss weights are fixed within each dataset.
Complete settings are reported in
Appendix~\ref{app:hyperparameter_settings}. Main test results are reported as
mean~\(\pm\)~standard deviation over three seeds. Constraint-solver
diagnostics and end-to-end inference cost are reported in
Appendices~\ref{app:projection_tolerance_runtime}
and~\ref{app:end_to_end_runtime}, respectively.

\paragraph{Metrics.}
Rollout \textbf{accuracy} is measured by \(\mathrm{RelL2}_r\), where \(r\) denotes the
rollout horizon. We quantify solver-level \textbf{feasibility} using
\(\LogDiv_r\), the base-10 logarithm of the discrete incompressibility
residual. On LC, we additionally report \(\LogBC_r\), defined analogously from the
\textbf{boundary-condition residual}. We additionally
report the \textbf{energy relative error }on P-NS and the \textbf{centerline profile error}
\(\mathcal E_{\mathrm{prof}}\) on LC. Formal definitions and the aggregation
protocol are provided in Appendix~\ref{app:metric_definitions}. 

\begin{table*}[t]
\centering
{\scriptsize
\setlength{\tabcolsep}{3pt}
\renewcommand{\arraystretch}{1.0}
\begin{tabular}{@{}llccccc@{}}
\toprule
Dataset
& Method
& \(\mathrm{RelL2}_{1}\downarrow\)
& \(\mathrm{RelL2}_{20}\downarrow\)
& \(\mathrm{RelL2}_{50}\downarrow\)
& \(\LogDiv_{50}\downarrow\)
& \(\LogBC_{50}\downarrow\) \\
\midrule
\multirow{6}{*}{P-NS}
& FNO
& $0.0044\pm0.0000$
& $5.0484\pm8.6469$
& $\nf$
& $\nf$
& \textemdash \\
& DivReg
& $0.0041\pm0.0000$
& $0.0290\pm0.0015$
& $0.0557\pm0.0080$
& $-2.0171\pm0.0089$
& \textemdash \\
& PINO
& $0.0047\pm0.0000$
& $0.0295\pm0.0003$
& $\underline{0.0477\pm0.0011}$
& $-1.9620\pm0.0067$
& \textemdash \\
& ClawNO
& $0.0038\pm0.0285$
& $\underline{0.0266\pm0.0015}$
& $0.9270\pm0.3536$
& $-5.3557\pm0.0179$
& \textemdash \\
& FinalProj
& ${0.0037\pm0.0001}$
& ${0.0281\pm0.0006}$
& $0.0503\pm0.0017$
& $\mathbf{-5.4981\pm0.0001}$
& \textemdash \\
& \PCMNO{}
& $\mathbf{0.0033\pm0.0001}$
& $\mathbf{0.0264\pm0.0002}$
& $\mathbf{0.0429\pm0.0005}$
& $\underline{-5.4978\pm0.0002}$
& \textemdash \\
\midrule
\multirow{5}{*}{LC}
& FNO
& $\underline{0.0042\pm0.0005}$
& $\underline{0.0049\pm0.0001}$
& $0.0058\pm0.0018$
& $-2.1344\pm0.0524$
& $-3.8770\pm0.0219$ \\
& DivReg
& $0.0167\pm0.0030$
& $0.0063\pm0.0017$
& $\underline{0.0050\pm0.0001}$
& $-3.0271\pm0.0128$
& $-3.9249\pm0.0160$ \\
& PINO
& $0.0433\pm0.0042$
& $0.2601\pm0.0142$
& $0.3305\pm0.0127$
& $-1.6027\pm0.0359$
& $-1.4254\pm0.0441$ \\
& FinalProj
& $0.0121\pm0.0062$
& $0.0123\pm0.0023$
& $0.0162\pm0.0019$
& $\underline{-4.0080\pm0.1363}$
& $\mathbf{-6.0000\pm0.0000}$ \\
& \PCMNO{}
& $\mathbf{0.0037\pm0.0001}$
& $\mathbf{0.0043\pm0.0002}$
& $\mathbf{0.0039\pm0.0001}$
& $\mathbf{-6.0625\pm0.2304}$
& $\mathbf{-6.0000\pm0.0000}$ \\
\bottomrule
\end{tabular}}
\caption{
In-distribution rollout accuracy and feasibility.
\(\LogDiv_{50}\) uses spectral divergence on P-NS and interior MAC control
volumes on LC; \(\LogBC_{50}\) is reported only for LC.
N.F. denotes a non-finite rollout before the stated horizon.
}
\label{tab:id_accuracy_feasibility}
\end{table*}

\begin{figure*}[t]
\centering
\includegraphics[width=0.7\textwidth]{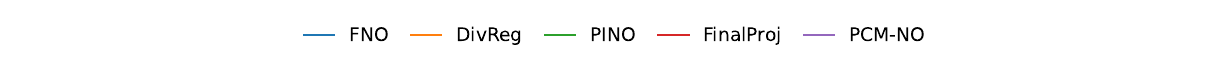}

\begin{minipage}[t]{0.25\textwidth}
\centering
\includegraphics[width=\linewidth]{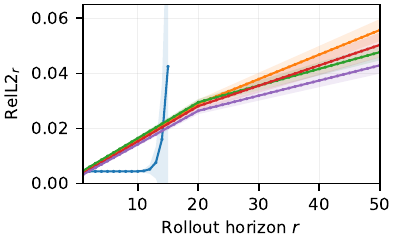}
\vspace{-0.3em}
{\scriptsize (a) P-NS rollout error}
\end{minipage}
\hspace{0.2cm}
\begin{minipage}[t]{0.25\textwidth}
\centering
\includegraphics[width=\linewidth]{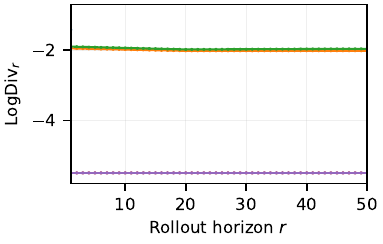}
\vspace{-0.3em}
{\scriptsize (b) P-NS divergence residual}
\end{minipage}
\hspace{0.2cm}
\begin{minipage}[t]{0.25\textwidth}
\centering
\includegraphics[width=\linewidth]{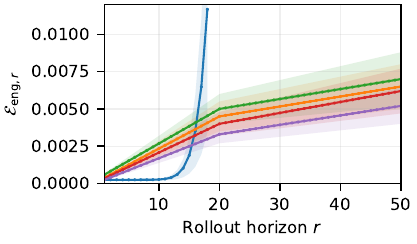}
\vspace{-0.3em}
{\scriptsize (c) P-NS energy error}
\end{minipage}

\vspace{0.55em}

\begin{minipage}[t]{0.25\textwidth}
\centering
\includegraphics[width=\linewidth]{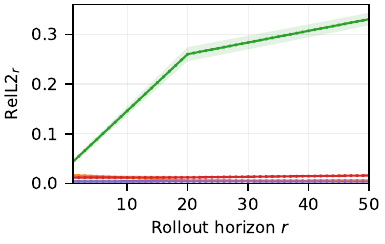}
\vspace{-0.3em}
{\scriptsize (d) LC rollout error}
\end{minipage}
\hspace{0.2cm}
\begin{minipage}[t]{0.25\textwidth}
\centering
\includegraphics[width=\linewidth]{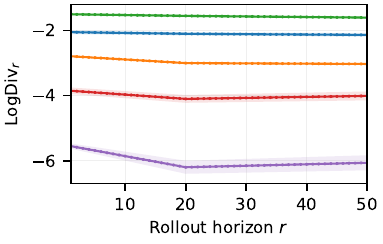}
\vspace{-0.3em}
{\scriptsize (e) LC divergence residual}
\end{minipage}
\hspace{0.2cm}
\begin{minipage}[t]{0.25\textwidth}
\centering
\includegraphics[width=\linewidth]{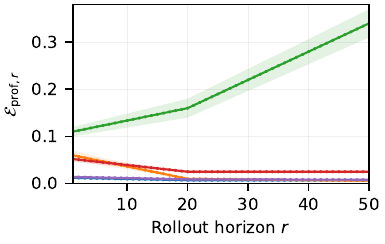}
\vspace{-0.3em}
{\scriptsize (f) LC centerline profile error}
\end{minipage}

\caption{
Long-rollout diagnostics. Shaded bands show the mean
\(\pm\) one standard deviation over three seeds. FNO curves are truncated
once the corresponding rollout becomes non-finite in the P-NS accuracy and
energy panels.
}
\label{fig:long_rollout_diagnostics}
\end{figure*}

\begin{table*}[t]
\centering
{\scriptsize
\setlength{\tabcolsep}{2.5pt}
\renewcommand{\arraystretch}{1.0}
\begin{tabular}{@{}llcccc@{}}
\toprule
Settings
& Method
& \(\mathrm{RelL2}_{50}\downarrow\)
& \(\LogDiv_{50}\downarrow\)
& \(\LogBC_{50}\downarrow\)
& \shortstack{Energy/Profile\\Err. \(\downarrow\)} \\
\midrule
\multirow{5}{*}{\shortstack[l]{P-NS-Re\\$Re\in\{700,1000\}$}}
& FNO
& \nf
& \nf
& \textemdash
& \nf \\
& DivReg
& $\underline{0.0999\pm0.0006}$
& $-1.9531\pm0.0079$
& \textemdash
&  $\underline{0.0091\pm0.0051}$ \\
& PINO
& $0.1091\pm0.0055$
& $-1.8891\pm0.0054$
& \textemdash
& $0.0105\pm0.0010$ \\
& FinalProj
& $0.1161\pm0.0103$
& $\underline{-5.5000\pm0.0009}$
& \textemdash
& $0.0275\pm0.0205$ \\
& \PCMNO{}
& $\mathbf{0.0964\pm0.0010}$
& $\mathbf{-5.5006\pm0.0005}$
& \textemdash
& $\mathbf{0.0059\pm0.0013}$ \\
\midrule
\multirow{5}{*}{\shortstack[l]{LC-Re\\$Re\in\{750,1000\}$}}
& FNO
& $\underline{0.0255\pm0.0050}$
& $-1.6761\pm0.0374$
& $-3.1378\pm0.1074$
& $\underline{0.1493\pm0.0257}$ \\
& DivReg
& $0.0373\pm0.0129$
& $-2.7939\pm0.0618$
& $-3.1870\pm0.0736$
& $0.2134\pm0.0387$ \\
& PINO
& $0.2799\pm0.0288$
& $-1.5863\pm0.1256$
& $-1.5279\pm0.0522$
& $0.8063\pm0.1889$ \\
& FinalProj
& $0.0550\pm0.0115$
& $\underline{-4.0374\pm0.1134}$
& $\mathbf{-6.0000\pm0.0000}$
& $0.3059\pm0.0762$ \\
& \PCMNO{}
& $\mathbf{0.0224\pm0.0082}$
& $\mathbf{-6.1617\pm0.2751}$
& $\mathbf{-6.0000\pm0.0000}$
& $\mathbf{0.1025\pm0.0324}$ \\
\midrule
\multirow{5}{*}{\shortstack[l]{LC-Lid\\$U_{\mathrm{lid}}=1.5$}}
& FNO
& $\underline{0.0596\pm0.0148}$
& $-1.3433\pm0.0264$
& $-2.4675\pm0.1289$
& $0.1415\pm0.0287$ \\
& DivReg
& $0.0839\pm0.0135$
& $-2.1814\pm0.0759$
& $-2.4631\pm0.0400$
& $0.1879\pm0.0155$ \\
& PINO
& $0.3579\pm0.0115$
& $-1.4067\pm0.0812$
& $-1.2386\pm0.0266$
& $0.6772\pm0.0122$ \\
& FinalProj
& $0.0601\pm0.0363$
& $\underline{-3.3582\pm0.1269}$
& $\mathbf{-6.0000\pm0.0000}$
& $\underline{0.1151\pm0.0432}$ \\
& \PCMNO{}
& $\mathbf{0.0143\pm0.0022}$
& $\mathbf{-5.9020\pm0.3165}$
& $\mathbf{-6.0000\pm0.0000}$
& $\mathbf{0.0311\pm0.0031}$ \\
\midrule
\multirow{5}{*}{\shortstack[l]{P-NS-Grid\\$64^2\!\to\!128^2$}}
& FNO
& \nf
& \nf
& \textemdash
& \nf \\
& DivReg
& $\underline{0.0511\pm0.0009}$
& $-1.0807\pm0.0588$
& \textemdash
& $0.0140\pm0.0033$ \\
& PINO
& $0.1002\pm0.0052$
& $-1.2159\pm0.0393$
& \textemdash
& $\underline{0.0080\pm0.0009}$ \\
& FinalProj
& $5.1918\mathrm{e}3\pm7.3833\mathrm{e}3$
& $\underline{-5.1451\pm0.0240}$
& \textemdash
& $1.1921\mathrm{e}{10}\pm2.0514\mathrm{e}{10}$ \\
& \PCMNO{}
& $\mathbf{0.0447\pm0.0004}$
& $\mathbf{-5.1798\pm0.0009}$
& \textemdash
& $\mathbf{0.0075\pm0.0007}$ \\
\bottomrule
\end{tabular}}
\caption{
Stress-set rollout results, using the in-distribution
validation-selected checkpoints without adaptation. Reynolds-number rows pool
the two test values. The last column reports energy error on P-NS and profile
error on LC; N.F. denotes a non-finite rollout.
}
\label{tab:stress_diagnostics}
\end{table*}

\subsection{In-Distribution Feasible Rollouts}
\label{sec:id_feasible_rollouts}

On P-NS, FinalProj, ClawNO, and \PCMNO{} maintain divergence-free rollouts near
the numerical floor of their constraint mechanisms, whereas the
soft-constraint methods remain near \(-2\) and FNO becomes non-finite before
\(50\) steps. \PCMNO{} achieves the lowest
\(\mathrm{RelL2}_{1}\), \(\mathrm{RelL2}_{20}\), and
\(\mathrm{RelL2}_{50}\). Although FinalProj reaches a nearly identical \(\LogDiv_{50}\), the shared
feasibility floor controls for output constraint satisfaction and isolates the
effect of transition parameterization. The lower rollout errors of \PCMNO{}
therefore support the empirical value of its projected-residual realization.
The corresponding tangent-filtering mechanism is examined directly in
Section~\ref{sec:mechanism_diagnostics}. Results in
Appendix~\ref{app:clawno_external_baseline} further show that \PCMNO{}
substantially improves accuracy and energy error over ClawNO, while the runtime
comparison quantifies the cost of the projection--retraction transition.

On LC, where \(C_h=[D_h;B_h]\), \PCMNO{} maintains both incompressibility and
the prescribed boundary values near the achieved solve floor while attaining
the lowest rollout errors. The joint
\(\LogDiv_{50}\)--\(\LogBC_{50}\) behavior shows that the relevant feasible set
is the complete affine constraint set and distinguishes hard solver-level
feasibility from soft residual reduction.

\subsection{Long-Horizon Rollout Stability}
\label{sec:long_rollout}

Figure~\ref{fig:long_rollout_diagnostics} traces accuracy, feasibility, and
flow-specific diagnostics across the rollout horizon. On P-NS, FinalProj and
\PCMNO{} maintain projection-level incompressibility, while the unconstrained
FNO rollout becomes non-finite. \PCMNO{} also gives the lowest long-horizon
state and energy errors among the finite methods. On LC, it keeps both Div and
BC near their numerical floors and substantially reduces the state and
centerline-profile errors. The results show that the feasible transition
remains effective under repeated composition and that its strongest predictive
advantage appears when incompressibility and prescribed boundary values must
be enforced jointly.

\subsection{Generalization and Stress Diagnostics}
\label{sec:stress_settings}

We evaluate four stress settings: P-NS Reynolds-number extrapolation
(\textbf{P-NS-Re}), LC Reynolds-number extrapolation (\textbf{LC-Re}), LC
lid-velocity shift (\textbf{LC-Lid}), and P-NS resolution transfer
(\textbf{P-NS-Grid}). Consistent with the pointwise closure property in
Proposition~\ref{prop:exact_closure}, Table~\ref{tab:stress_diagnostics} shows
that \PCMNO{} preserves near-numerical-floor feasibility beyond the training
distribution. It maintains \(\LogDiv_{50}\) on P-NS-Re, both
\(\LogDiv_{50}\) and \(\LogBC_{50}\) on LC-Re and LC-Lid, and
near-feasibility after reconstructing the constraint operators on the
\(128^2\) P-NS grid. Across all four settings, \PCMNO{} also achieves the
lowest \(\mathrm{RelL2}_{50}\) and flow-specific diagnostic error, whereas
FinalProj reaches a comparable feasibility floor, so the lower state and
flow-diagnostic errors of \PCMNO{} isolate the predictive advantage of the
projected-residual realization under distribution shift. LC-Lid directly changes the affine target \(b_h\), and P-NS-Grid
evaluates target-grid transfer under reconstructed constraint operators in a
setting where grid transfer, unseen-frequency extrapolation, and aliasing are
closely coupled \citep{sakarvadia2026false}.

\begin{figure*}[t]
\centering
\includegraphics[width=0.8\textwidth]
{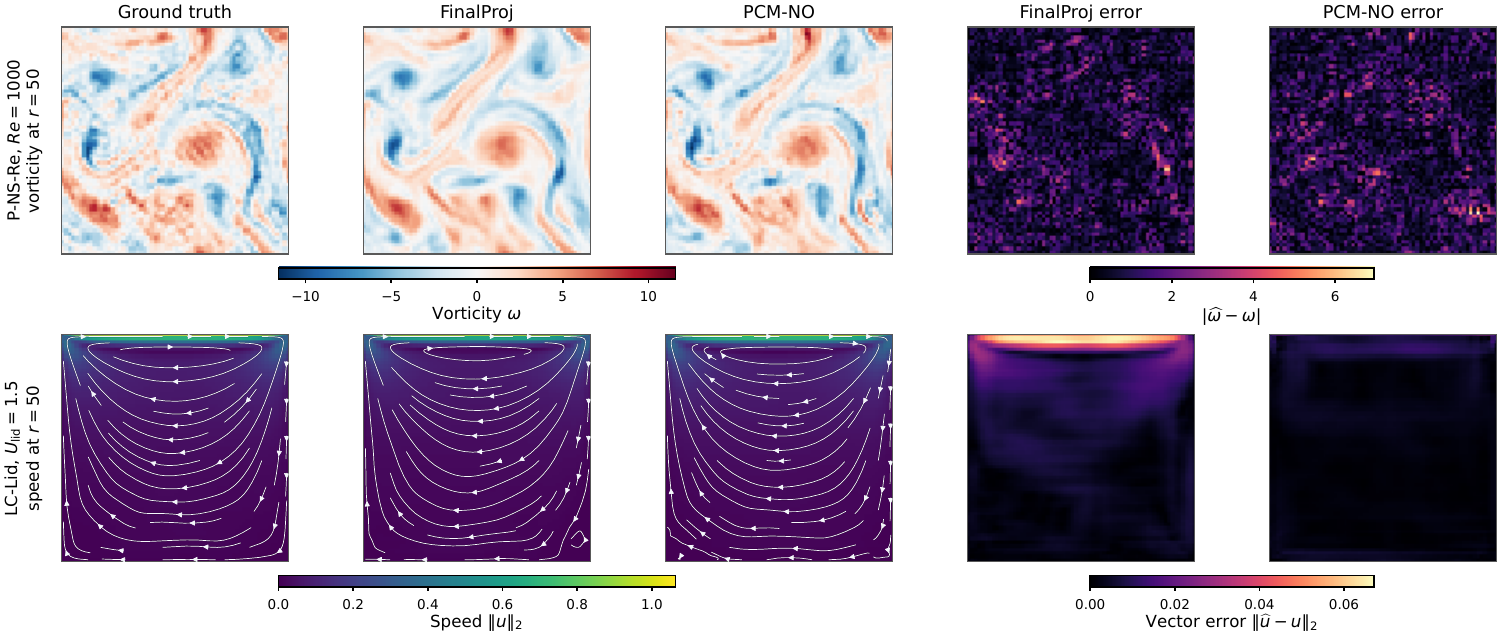}
\caption{
Representative \(50\)-step stress predictions: P-NS Reynolds extrapolation
at \(Re=1000\) and the LC lid shift at \(U_{\mathrm{lid}}=1.5\).
Rows share field and error color scales, respectively. Trajectories are
selected by median ground-truth kinetic energy, independently of model error.
}
\label{fig:qualitative_fields}
\end{figure*}

Figure~\ref{fig:qualitative_fields} further shows that \PCMNO{} better
preserves the P-NS vorticity organization and LC cavity circulation after
\(50\) steps, consistent with Table~\ref{tab:stress_diagnostics}. Detailed
protocols and metric definitions are provided in
Appendices~\ref{app:constraint_configurations_stress_protocol}
and~\ref{app:metric_definitions}.

\subsection{Mechanism Diagnostics}
\label{sec:mechanism_diagnostics}

\paragraph{Tangent filtering.}
Proposition~\ref{prop:tangent_filtering} states that projection onto the
discrete tangent space cannot increase the distance from a candidate update to
an admissible target. We test this property on realized updates from the trained
\PCMNO{} models by comparing raw and projected candidate updates with the
projected reference increment. As shown in
Table~\ref{tab:tangent_filtering}, projection consistently reduces this error
on both datasets, while the normalized Pythagorean gap remains near numerical
precision. The removed normal component also accounts for a substantial
fraction of the candidate-update energy, confirming that the tangent layer
actively filters constraint-incompatible directions. This diagnostic verifies
the local mechanism in Proposition~\ref{prop:tangent_filtering}, while
Section~\ref{sec:component_ablations} evaluates its role in the complete
transition.

\begin{figure}[t]
    \centering
    \begin{subfigure}[t]{0.22\textwidth}
        \centering
        \includegraphics[width=\linewidth]
        {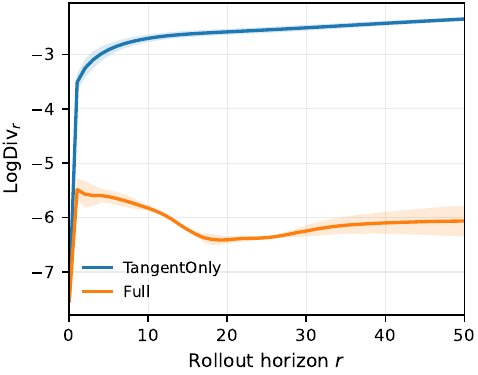}
        \caption{Accumulation versus per-step reset.}
        \label{fig:residual_reset_accumulation}
    \end{subfigure}
    \hfill
    \begin{subfigure}[t]{0.22\textwidth}
        \centering
        \includegraphics[width=\linewidth]
        {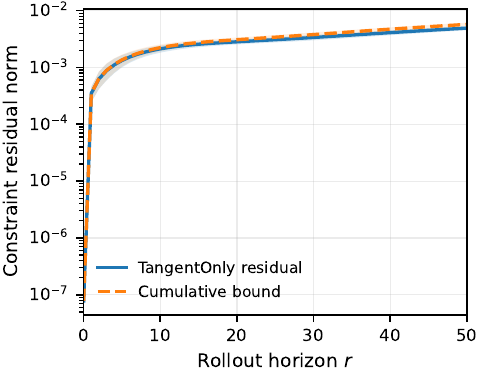}
        \caption{Realized residual and cumulative upper bound.}
        \label{fig:residual_reset_bound}
    \end{subfigure}
    \caption{
Controlled LC replay with identical projected updates.
}
    \label{fig:residual_reset}
\end{figure}

\paragraph{Residual accumulation and reset.}
Figure~\ref{fig:residual_reset} examines the behavior predicted by
Proposition~\ref{prop:approximate_residuals}. In a controlled LC replay,
candidate updates generated along the full \PCMNO{} rollout are projected once
using the evaluation solve budget and applied to two branches. TangentOnly
omits the final retraction, whereas Full retracts the same update sequence after
every step, so the branches differ only in the state-level reset. TangentOnly
accumulates incompressibility error, while Full remains near the achieved
constraint-solve floor. Both preserve the prescribed boundary values because
the projected updates satisfy the homogeneous boundary constraint, and the
tangent-only residual remains below the cumulative projection-residual bound.
These results directly support the accumulation-versus-reset interpretation in
Proposition~\ref{prop:approximate_residuals}.

\begin{table}[t]
\centering
{\scriptsize
\setlength{\tabcolsep}{2.5pt}
\renewcommand{\arraystretch}{1.0}
\begin{tabular}{@{}lcccc@{}}
\toprule
Dataset
& \shortstack{Raw\\RelErr. \(\downarrow\)}
& \shortstack{Projected\\RelErr. \(\downarrow\)}
& \shortstack{Normal\\Frac. (\%)}
& \shortstack{Pythag.\\Gap \(\downarrow\)} \\
\midrule
P-NS
& \(0.8220 \pm 0.0258\)
& \(0.0607 \pm 0.0014\)
& \(39.75 \pm 1.52\)
& \(\left(5.22 \pm 0.16\right)\times10^{-8}\) \\
LC
& \(3.8898 \pm 0.8964\)
& \(0.0559 \pm 0.0090\)
& \(87.76 \pm 3.19\)
& \(\left(5.96 \pm 0.37\right)\times10^{-8}\) \\
\bottomrule
\end{tabular}}
\caption{
Tangent-filtering diagnostic.
Errors are measured against the projected reference increment;
Normal Frac. is the removed candidate-update energy fraction, and
Pythag. Gap is the normalized residual of Proposition~\ref{prop:tangent_filtering}.
}
\label{tab:tangent_filtering}
\end{table}

\subsection{Component Ablations}
\label{sec:component_ablations}

Having directly verified tangent filtering and residual reset, we next examine
how tangent projection, state retraction, and the complete affine constraint
affect the learned rollout. We conduct the ablation on LC because it activates
both incompressibility and prescribed boundary constraints.
We consider three structural variants. \textbf{TangentOnly} retains the
residual-update parameterization and the tangent projection, but omits the
final retraction:
$
u_{h,k+1}
=
u_{h,k}
+
\Delta t\,P_hK_\theta(u_{h,k},\mu,g_h,h).
$
\textbf{RetractOnly} removes the tangent projection and retracts the
unconstrained residual update:
$
u_{h,k+1}
=
R_h\!\left(
u_{h,k}
+
\Delta t\,K_\theta(u_{h,k},\mu,g_h,h);
b_h
\right).
$
\textbf{DivOnly} applies both fixed layers using the reduced constraint
\(C_h=D_h\), thereby removing the prescribed boundary constraint
\(B_hu_h=g_h\). The full \PCMNO{} transition retains tangent projection,
boundary-aware constraints, and state retraction.

Table~\ref{tab:lc_ablation} separates three numerical roles.
TangentOnly retains the projected residual update but removes the state-level
reset. Its increasing constraint residual is consistent with the accumulation
mechanism in Proposition~\ref{prop:approximate_residuals}. RetractOnly applies the reset to an unconstrained residual update and therefore
recovers much of the feasibility and accuracy. Relative to RetractOnly,
\PCMNO{} reduces \(\mathrm{RelL2}_{50}\) by \(36\%\) and
\(\mathcal E_{\mathrm{prof},50}\) by \(58\%\), while improving
\(\LogDiv_{50}\) by \(3.35\) log units. These gains quantify the empirical
benefit of the explicit projected-update parameterization under finite
constraint solves. DivOnly removes the
prescribed-boundary component from \(C_h\), leading to substantial boundary
violation and degraded profile accuracy. Together, the results support the complete affine constraint, the per-step
feasibility reset, and the explicit admissible-update parameterization as
complementary components of the implemented transition.

\paragraph{Computational implications.}
To place the accuracy--feasibility gains in a solver context,
Appendix~\ref{app:end_to_end_runtime} compares complete learned transitions
with the corresponding reference solvers over the same physical horizon.
At batch size one, \PCMNO{} remains \(7.6\times\) faster than the P-NS
reference solver and \(6.4\times\) faster than the LC reference solver over a
\(50\)-step rollout. Constraint enforcement costs \(1.37\times\) the FNO
rollout on P-NS, but \(41.53\times\) on LC, where the two iterative constraint
solves dominate inference. The computational overhead of the solver-defined
constraint layers is therefore discretization-dependent, motivating warm
starts, preconditioning, and adaptive solve budgets for boundary-aware
settings.

\begin{table}[t]
\centering
{\scriptsize
\setlength{\tabcolsep}{3.6pt}
\renewcommand{\arraystretch}{1.0}
\begin{tabular}{@{}lcccc@{}}
\toprule
Variant
& \(\mathrm{RelL2}_{50}\downarrow\)
& \(\LogDiv_{50}\downarrow\)
& \(\LogBC_{50}\downarrow\)
& \(\mathcal E_{\mathrm{prof},50}\downarrow\) \\
\midrule
TangentOnly
& $0.0156\pm0.0021$
& $-1.2424\pm0.0587$
& $\mathbf{-6.0000\pm0.0000}$
& $0.0723\pm0.0129$ \\

RetractOnly
& $\underline{0.0061\pm0.0001}$
& $-2.7142\pm0.0288$
& $\mathbf{-6.0000\pm0.0000}$
& $\underline{0.0208\pm0.0007}$ \\

DivOnly
& $0.0739\pm0.0121$
& $\underline{-2.7606\pm0.1050}$
& $-1.8536\pm0.0653$
& $0.2075\pm0.0269$ \\

\PCMNO{}
& $\mathbf{0.0039\pm0.0001}$
& $\mathbf{-6.0625\pm0.2304}$
& $\mathbf{-6.0000\pm0.0000}$
& $\mathbf{0.0088\pm0.0027}$ \\
\bottomrule
\end{tabular}}
\caption{
LC component ablation with \(50\)-step rollouts.
}
\label{tab:lc_ablation}
\end{table}

\section{Related Work}

\paragraph{Neural operators and temporal PDE surrogates.}
Neural operators learn mappings between function spaces through DeepONet,
FNO, and their factorized, geometry-adaptive, convolutional, localized,
mesh-based, graph, and transformer variants
\citep{lu2021learning,li2021fourier,kovachki2023neural,
tran2023factorized,li2023geofno,raonic2023convolutional,
liuschiaffini2024localized,pfaff2021learning,hao2023gnot,
wu2024transolver}.
Foundation models and benchmarks broaden the task scope
\citep{herde2024poseidon,takamoto2022pdebench}, while autoregressive,
refinement, memory, and stability-oriented methods target long-horizon
accuracy
\citep{brandstetter2022message,mccabe2023towards,lippe2023pde,
ruiz2025memory}.
These works do not generally require every composed transition to remain in a
solver-defined feasible set.

\paragraph{Physics-aware and constrained dynamics.}
Physics-informed, conservation aware, divergence-free, and boundary-aware
operators incorporate PDE structure through residual objectives, data-free or
boundary-integral formulations, and specialized parameterizations
\citep{raissi2019physics,wang2021physicsdeeponet,li2022physics,
karniadakis2021physics,choi2025datafree,fang2023boundaries,
liu2024clawno,liu2023ino,richterpowell2022neural,
espath2021statistical,cheng2026decoupled,wang2024beno}.
EquiNO imposes equilibrium through divergence-free POD bases, while CompNO
composes learned convection and diffusion modules with a Poisson solver
\citep{eivazi2025equino,hmida2026compno}.
Related structure-preserving methods use Hodge decompositions, conservative
transport, port--Hamiltonian factorizations, hard-constrained generative
sampling, or classical projection methods
\citep{zheng2026hodge,lan2026fluxnet,noguchi2026mesh,
utkarsh2025physics,chorin1968numerical,temam1969sur,
guermond2006overview,arnold2006finite}.
Neural--numerical and explicitly constrained models similarly combine learned
mappings with analytical or iterative structure
\citep{greydanus2019hamiltonian,finzi2020explicit,
rudikov2024fcgno,hao2024newton}, and post-hoc projection can enforce
divergence-free outputs \citep{li2026project}.
\PCMNO{} instead defines an autoregressive map between solver-defined affine
feasible sets through a projected residual parameterization, supporting joint
constraints and nonhomogeneous targets while enabling analysis of exact
closure and approximate-solve residual propagation.

\section{Conclusion}

We introduced \PCMNO{}, an autoregressive neural-operator construction whose
transition is closed on a solver-defined affine constraint manifold. The backbone predicts a candidate update direction, while fixed tangent projection
and state-retraction layers determine the admissible increment and the
feasible output state. The structural analysis establishes exact
feasible-state closure, orthogonal filtering of constraint-incompatible
increment components, and residual accumulation and reset bounds for
approximate numerical solves. Across periodic and boundary-constrained
incompressible-flow settings, \PCMNO{} maintains discrete incompressibility
and prescribed boundary values over long rollouts and under controlled
parameter, boundary-value, and resolution shifts. The experiments also show
consistent improvements in rollout accuracy and flow-specific diagnostics.
Extending the construction to nonlinear, state-dependent, and
geometry-dependent constraint sets is a natural direction for future work.

\section{Limitations and Ethical Considerations}

The current formulation is limited to fixed linear discrete constraints and
nonempty affine feasible sets. The experiments cover two-dimensional
incompressible flows with an FNO backbone, while three-dimensional systems,
irregular geometries, and nonlinear or state-dependent constraints remain to
be studied. The cost of the constraint layers is also discretization
dependent, with the iterative LC solver introducing substantially greater
overhead than the spectral P-NS implementation.

The datasets are synthetic and contain no human or personally identifiable
information. Learned CFD surrogates may nevertheless produce plausible but
inaccurate predictions under untested conditions. Safety-critical use therefore
requires independent validation, error estimation, and a verified numerical
solver as fallback.

\section{Generative AI Usage}

Generative AI tools were used to assist language editing, structural revision,
and code drafting. The authors reviewed all generated content and independently
verified the mathematical derivations, implementations, experimental
protocols, and reported results. Generative AI was not treated as an author and
did not make autonomous scientific decisions.

\bibliographystyle{ACM-Reference-Format}
\bibliography{pcm_no_references}

\appendix

\section{Proofs and Approximate Constraint Solves}
\label{app:proofs}

This appendix proves the structural statements in Section~\ref{sec:theory} and relates approximate constraint layers to the achieved residuals of their linear systems.

\subsection{Proof of Proposition~\ref{prop:exact_closure}}
\label{app:proof_exact_closure}

Let
\[
    A_h
    =
    C_hW_h^{-1}C_h^\top.
\]
For every \(z\in\mathbb R^{m_h}\),
\[
    z^\top A_h z
    =
    \|
        W_h^{-1/2}C_h^\top z
    \|_2^2.
\]
Hence
\(
\ker A_h=\ker C_h^\top
\).
Because \(A_h\) is symmetric,
\[
    \operatorname{range}(A_h)
    =
    (\ker A_h)^\perp
    =
    (\ker C_h^\top)^\perp
    =
    \operatorname{range}(C_h).
\]

The KKT conditions for the tangent projection yield
\[
    P_hv_h
    =
    v_h
    -
    W_h^{-1}C_h^\top
    A_h^\dagger C_hv_h.
\]
Since \(C_hv_h\in\operatorname{range}(C_h)
=\operatorname{range}(A_h)\),
\[
\begin{aligned}
    C_hP_hv_h
    &=
    C_hv_h
    -
    A_hA_h^\dagger C_hv_h
    \\
    &=0.
\end{aligned}
\]

For the state projection,
\[
    R_h(w_h;b_h)
    =
    w_h
    -
    W_h^{-1}C_h^\top
    A_h^\dagger(C_hw_h-b_h).
\]
The nonemptiness of \(\mathcal M_h(b_h)\) implies
\(b_h\in\operatorname{range}(C_h)\), and therefore
\(
C_hw_h-b_h\in\operatorname{range}(A_h)
\).
It follows that
\[
\begin{aligned}
    C_hR_h(w_h;b_h)
    &=
    C_hw_h
    -
    A_hA_h^\dagger(C_hw_h-b_h)
    \\
    &=b_h.
\end{aligned}
\]
Finally, if \(u_{h,k}\in\mathcal M_h(b_h)\), then
\[
\begin{aligned}
    C_h
    \left(
        u_{h,k}
        +
        \Delta t\,
        P_hK_\theta(u_{h,k},\mu,g_h,h)
    \right)
    &=
    C_hu_{h,k}
    +
    \Delta t\,
    C_hP_hK_\theta(u_{h,k},\mu,g_h,h)
    \\
    &=b_h.
\end{aligned}
\]
The subsequent state projection preserves membership in
\(\mathcal M_h(b_h)\). Induction over \(k\) proves the rollout
statement.

\subsection{Derivation of Eq.~\eqref{eq:posthoc_instance}}
\label{app:exact_affine_relation}

Define
\[
    Q_h
    =
    W_h^{-1}C_h^\top A_h^\dagger,
    \qquad
    P_h=I-Q_hC_h .
\]
For \(u_h\in\mathcal M_h(b_h)\) and any feasible target value \(b_h'\),
\begin{align*}
    R_h(u_h+\Delta t\,v_h;b_h')
    &=
    u_h+\Delta t\,v_h
    -Q_h\bigl(b_h+\Delta t\,C_hv_h-b_h'\bigr)
    \\
    &=
    u_h+\Delta t\,P_hv_h-Q_h(b_h-b_h')
    \\
    &=
    R_h(u_h+\Delta t\,P_hv_h;b_h').
\end{align*}
Taking \(b_h'=b_h\) gives
\[
    R_h(u_h+\Delta t\,v_h;b_h)
    =
    u_h+\Delta t\,P_hv_h,
\]
which proves Eq.~\eqref{eq:posthoc_instance}.

\subsection{Proof of Proposition~\ref{prop:tangent_filtering}}
\label{app:proof_tangent_filtering}

The projection \(P_h\) is the \(W_h\)-orthogonal projection onto
\(\mathcal U_h=\ker C_h\). Hence
\(P_h\widetilde{\delta u}_h\in\mathcal U_h\), while
\((I-P_h)\widetilde{\delta u}_h\) is \(W_h\)-orthogonal to
\(\mathcal U_h\). Since \(\delta u_h^\star\in\mathcal U_h\),
\[
    \left\langle
        P_h\widetilde{\delta u}_h-\delta u_h^\star,
        (I-P_h)\widetilde{\delta u}_h
    \right\rangle_{W_h}
    =0 .
\]
Therefore, the decomposition
\[
    \widetilde{\delta u}_h-\delta u_h^\star
    =
    \bigl(
        P_h\widetilde{\delta u}_h-\delta u_h^\star
    \bigr)
    +
    (I-P_h)\widetilde{\delta u}_h
\]
has \(W_h\)-orthogonal components. Taking squared \(W_h\)-norms
gives the Pythagorean identity in
Proposition~\ref{prop:tangent_filtering}. Dropping the nonnegative
normal-component term yields the stated inequality.

\subsection{Approximate Solves and Proof of
Proposition~\ref{prop:approximate_residuals}}
\label{app:proof_residual_reset}

Let
\[
    A_h=C_hW_h^{-1}C_h^\top.
\]
The implementation may solve a damped normal equation
\[
    (A_h+\gamma_h I)\widehat\lambda=d_h
\]
with \(\gamma_h\ge0\). This includes the undamped case
\(\gamma_h=0\).

For tangent projection, set \(d^P_{h,k}=C_hv_{h,k}\), define the
achieved linear-system residual by
\[
    \rho^P_{h,k}
    =
    (A_h+\gamma_h I)\widehat\lambda^P_{h,k}
    -
    d^P_{h,k},
\]
and let
\[
    \widehat P_{h,k}v_{h,k}
    =
    v_{h,k}
    -
    W_h^{-1}C_h^\top\widehat\lambda^P_{h,k}.
\]
Then
\begin{align}
    C_h\widehat P_{h,k}v_{h,k}
    &=
    d^P_{h,k}
    -
    A_h\widehat\lambda^P_{h,k}
    \\
    &=
    \gamma_h\widehat\lambda^P_{h,k}
    -
    \rho^P_{h,k}.
\end{align}
Thus one may take
\[
    \epsilon^P_{h,k}
    =
    \|
        \gamma_h\widehat\lambda^P_{h,k}
        -
        \rho^P_{h,k}
    \|,
\]
or use the upper bound
\[
    \epsilon^P_{h,k}
    \le
    \|
        \rho^P_{h,k}
    \|
    +
    \gamma_h
    \|
        \widehat\lambda^P_{h,k}
    \|.
\]

For state projection, define
\[
    d^R_{h,k}
    =
    C_hw_h-b_{h,k+1},
\]
\[
    \rho^R_{h,k}
    =
    (A_h+\gamma_h I)\widehat\lambda^R_{h,k}
    -
    d^R_{h,k},
\]
and
\[
    \widehat R_{h,k}(w_h;b_{h,k+1})
    =
    w_h
    -
    W_h^{-1}C_h^\top\widehat\lambda^R_{h,k}.
\]
The resulting feasibility residual is
\[
\begin{aligned}
    C_h\widehat R_{h,k}(w_h;b_{h,k+1})
    -
    b_{h,k+1}
    &=
    \gamma_h\widehat\lambda^R_{h,k}
    -
    \rho^R_{h,k}.
\end{aligned}
\]
Consequently,
\[
    \epsilon^R_{h,k}
    =
    \|
        \gamma_h\widehat\lambda^R_{h,k}
        -
        \rho^R_{h,k}
    \|
\]
is valid. When \(\gamma_h=0\), these expressions reduce to the
undamped residual identities.

For the tangent-only rollout,
\[
    r^{\mathrm{tan}}_{h,k+1}
    =
    r^{\mathrm{tan}}_{h,k}
    +
    \Delta t\,
    C_h\widehat P_{h,k}v_{h,k}.
\]
Repeated application of the triangle inequality gives
\[
    \|
        r^{\mathrm{tan}}_{h,k}
    \|
    \le
    \|
        r^{\mathrm{tan}}_{h,0}
    \|
    +
    \Delta t
    \sum_{j=0}^{k-1}
    \epsilon^P_{h,j},
\]
which proves Eq.~\eqref{eq:tangent_residual_bound}. The full and
retract-only transitions end each step with
\(\widehat R_{h,k}\), so
Eq.~\eqref{eq:retraction_reset_bound} follows directly from
Assumption~\ref{ass:approximate_layers}.

\section{Discrete Constraint Operators}
\label{app:discrete_constraint_operators}

This appendix specifies the discrete operators used in the two constraint
configurations. The same operators are used for data generation, projection or
retraction, training-time rollout, and evaluation.

\subsection{P-NS Spectral Constraint}
\label{app:pns_constraint}

For P-NS, \(C_h=D_h\) is the periodic spectral divergence operator.
Let \(G_h\) denote the corresponding spectral gradient. Under the
discrete periodic inner product,
\[
    G_h=-D_h^{*_{W_h}},
\]
so the Helmholtz projection
\begin{equation}
    P_h
    =
    I
    -
    G_h(D_hG_h)^\dagger D_h
\end{equation}
is the \(W_h\)-orthogonal projection onto \(\ker D_h\). It is
implemented in Fourier space, with the zero mode left unchanged. The same
spectral divergence operator is used by the constraint layer and by the
reported P-NS feasibility metric.

\subsection{LC MAC-Grid Constraint}
\label{app:lc_constraint}

LC states are represented on a MAC-style finite-volume grid. Let
\(B_hu_h=g_h\) encode the no-slip walls and the moving lid, and let
\(D_hu_h=0\) denote finite-volume incompressibility on the interior
control volumes. Formally,
\[
    C_h
    =
    \begin{bmatrix}
        D_h\\
        B_h
    \end{bmatrix},
    \qquad
    b_h
    =
    \begin{bmatrix}
        0\\
        g_h
    \end{bmatrix}.
\]

The implementation enforces this stacked constraint through boundary lifting
and a reduced-coordinate projection. Let
\(\ell_h(g_h)\) be the stored MAC field containing the prescribed wall and
ghost values, and let
\[
    E_h:\mathcal Z_h\to\mathcal X_h
\]
inject free-face variables into the full MAC state. Every state satisfying the
boundary constraint has the unique representation
\begin{equation}
    u_h
    =
    \ell_h(g_h)
    +
    E_hz_h,
    \qquad
    B_hE_h=0,
    \qquad
    B_h\ell_h(g_h)=g_h.
    \label{eq:lc_reduced_coordinates}
\end{equation}
Define
\[
    \overline D_h
    =
    D_hE_h,
    \qquad
    c_h(g_h)
    =
    -D_h\ell_h(g_h).
\]
Then
\[
    D_hu_h=0
    \quad\Longleftrightarrow\quad
    \overline D_hz_h=c_h(g_h).
\]
Thus the full constraint \(C_hu_h=b_h\) is equivalent to a reduced
divergence constraint on the free velocity faces.

Let
\[
    \overline W_h
    =
    E_h^\top W_hE_h.
\]
The reduced tangent projection is
\begin{equation}
    \overline P_h
    =
    I
    -
    \overline W_h^{-1}\overline D_h^\top
    \left(
        \overline D_h
        \overline W_h^{-1}
        \overline D_h^\top
    \right)^\dagger
    \overline D_h,
    \label{eq:lc_reduced_tangent_projection}
\end{equation}
and the reduced state projection is
\begin{equation}
\begin{aligned}
    \overline R_h(z_h;c_h)
    &=
    z_h
    -
    \overline W_h^{-1}\overline D_h^\top
    \left(
        \overline D_h
        \overline W_h^{-1}
        \overline D_h^\top
    \right)^\dagger
    \left(
        \overline D_hz_h-c_h
    \right).
\end{aligned}
\label{eq:lc_reduced_state_projection}
\end{equation}

The current MAC implementation uses the Euclidean metric on the free faces,
so \(\overline W_h=I\). Candidate increments are first assigned homogeneous
boundary values and then projected with
Eq.~\eqref{eq:lc_reduced_tangent_projection}. State retraction first applies
the boundary lifting and then solves
Eq.~\eqref{eq:lc_reduced_state_projection}. The boundary constraint is
therefore imposed by construction, while the remaining divergence residual is
controlled by the reduced normal-equation solve.

The exact formulas correspond to a converged undamped solve. The implemented
finite-iteration CG solver uses a small damping parameter, and its residual
behavior is covered by Proposition~\ref{prop:approximate_residuals} and
Appendix~\ref{app:proof_residual_reset}. The Div values reported for LC are
evaluated on the interior MAC control volumes.

\section{Experimental Details}
\label{app:experimental_details}

\subsection{Dataset Generation and Statistics}
\label{app:dataset_generation}

P-NS is generated as forced periodic incompressible flow on
\([0,2\pi]^2\) with \(Re\in[100,500]\). LC is generated on \([0,1]^2\)
using a MAC-grid projection solver with no-slip walls and
\(U_{\mathrm{lid}}\in[0.8,1.2]\). Within each dataset, all methods use the same
training, validation, and test trajectories. Table~\ref{tab:dataset_statistics}
summarizes the data generation, discretization, and evaluation settings.

\begin{table}[t]
\centering
{\scriptsize
\setlength{\tabcolsep}{3.0pt}
\renewcommand{\arraystretch}{1.02}
\begin{tabular}{@{}lcc@{}}
\toprule
Setting & P-NS & LC \\
\midrule
Domain
& \([0,2\pi]^2\)
& \([0,1]^2\) \\

Generator
& \shortstack{Dealiased pseudo-spectral\\RK4 solver}
& \shortstack{MAC-grid projection\\solver} \\

State representation
& Velocity field
& Staggered velocity field \\

Spatial resolution
& \(64\times64\)
& \(64\times64\) \\

Stored states per trajectory
& \(56\)
& \(56\) \\

One-step transitions per trajectory
& \(55\)
& \(55\) \\

Saved-state interval
& \(0.05\)
& \(0.006\) \\

Burn-in
& \(8\) saved intervals
& \(4\) saved intervals \\

Train/validation/test trajectories
& \(500/100/200\)
& \(500/100/200\) \\

Training Reynolds numbers
& \(Re\in[100,500]\)
& \(Re\in[100,500]\) \\

Training boundary data
& Periodic
& \(U_{\mathrm{lid}}\in[0.8,1.2]\) \\

Training rollout horizon
& \(H=4\)
& \(H=4\) \\

ID evaluation horizons
& \(r\in\{1,20,50\}\)
& \(r\in\{1,20,50\}\) \\
\bottomrule
\end{tabular}}
\caption{
Dataset generation and evaluation statistics. Spatial resolution refers to the
active computational grid, and stored states include the initial state of each
trajectory. All methods use identical trajectory splits within each dataset.
}
\label{tab:dataset_statistics}
\end{table}

\subsection{Constraint configurations and stress protocol}
\label{app:constraint_configurations_stress_protocol}

The experiments use two constraint configurations as the organizing axis for
the empirical study. This organization follows the central claim of \PCMNO{}:
the learned autoregressive transition should map a solver-defined feasible set
to itself. Both datasets are generated from two-dimensional incompressible
Navier--Stokes dynamics, but they induce different affine constraint sets. This
separation allows us to distinguish divergence preservation from
boundary-aware feasibility. The continuous equations below define the
data-generating families, while the reported feasibility metrics and the
projection--retraction layers are defined using the corresponding discrete
solver operators.

\paragraph{P-NS}
P-NS denotes periodic Navier--Stokes, instantiated as forced periodic
Kolmogorov flow. Let \(\Omega=\mathbb T^2_{2\pi}\) denote the two-dimensional torus. In
continuous form, the P-NS data-generating family is
\[
\begin{aligned}
    \partial_t u(x,t) + u(x,t)\cdot\nabla u(x,t)
    &= -\nabla p(x,t) + \nu \Delta u(x,t) + f_{\mathrm K}(x), \\
    \nabla\cdot u(x,t) &= 0,
\end{aligned}
\]
where $(x,t)\in \Omega\times(0,T]$. Equivalently, when represented on the periodic box \([0,2\pi)^2\), the
velocity field satisfies
\[
    u(x+2\pi e_i,t)=u(x,t),\qquad i=1,2 .
\]
Here \(u=(u_x,u_y)\) is the velocity field, \(p\) is the pressure,
\(\nu=1/Re\) in our nondimensionalized data generation, and
\(f_{\mathrm K}\) is a fixed Kolmogorov forcing. At the discrete level, the
feasible set is
\[
    \mathcal M_h(0)=\{u_h:D_hu_h=0\},
\]
so the only explicit constraint operator is the periodic spectral divergence
\(D_h\). Periodicity is built into the discretization, and no separate BC
residual is reported. This configuration provides a clean test of
divergence-free rollout, energy-based diagnostics, Reynolds-number
extrapolation, and resolution transfer because the spectral projection and the
energy diagnostic are defined on the same periodic grid.

\paragraph{LC}
LC denotes lid-driven cavity flow. The \emph{lid} is the top wall of the
unit-square cavity. The left, right, and bottom walls are stationary no-slip
walls, while the top lid moves horizontally with prescribed speed
\(U_{\mathrm{lid}}\). The continuous data-generating equation is
\[
\begin{aligned}
    \partial_t u + u\cdot\nabla u
    &= -\nabla p + \nu \Delta u, \\
    \nabla\cdot u &= 0,
\end{aligned}
\qquad
x\in[0,1]^2,\quad t>0,
\]
with boundary condition $u=(0,0)$ on the left, right, and bottom walls, $ u=(U_{\mathrm{lid}},0)$ on the top lid.
The resulting discrete feasible set is
\[
    \mathcal M_h(b_h)=
    \{u_h:D_hu_h=0,\;B_hu_h=g_h\},
    \qquad
    b_h=\begin{bmatrix}0\\ g_h\end{bmatrix}.
\]
Equivalently, \(C_h=[D_h;B_h]\) and \(C_hu_h=b_h\). This setting is the main
test of the general constraint formulation. It requires a rollout to preserve
both incompressibility and prescribed boundary values, and it prevents the
method from being interpreted as only a periodic Helmholtz projection. The
profile error reported for LC uses the vertical and horizontal centerline
velocity profiles, which are more informative for cavity flow than a
spectral-energy diagnostic.

\paragraph{Evaluation-only stress diagnostics.}
The main stress diagnostics reuse the validation-selected checkpoints from the
in-distribution experiments. Test data are regenerated under controlled changes
in the physical or numerical condition, while the trained model, normalization
statistics, backbone hyperparameters, and soft-loss weights are kept fixed. No
stress-set tuning or instance-wise test-time fine-tuning is used. This protocol
makes the stress results a direct test of out-of-distribution rollout behavior
without additional adaptation.

\paragraph{Reynolds-number extrapolation.}
For P-NS, the training range is \(Re\in[100,500]\) and the stress test uses
\(Re\in\{700,1000\}\). For LC, the stress test uses
\(Re\in\{750,1000\}\), while the lid velocity remains in the training range
\(U_{\mathrm{lid}}\in[0.8,1.2]\). Keeping the LC lid distribution fixed isolates
the effect of Reynolds-number extrapolation from a boundary-value shift. Since
higher Reynolds numbers correspond to smaller viscosity and stronger
small-scale or boundary-layer effects, these tests probe whether the learned
transition remains stable beyond the physical-parameter range seen during
training.

\paragraph{Boundary-value shift on LC}
The lid-shift experiment keeps \(Re\in[100,500]\) and evaluates at
\(U_{\mathrm{lid}}=1.5\), outside the training range. This condition changes the
right-hand side of the constraint system through \(g_h(U_{\mathrm{lid}})\) and
therefore changes the feasible set \(\mathcal M_h(b_h)\) itself. It directly
tests whether boundary lifting and boundary-aware retraction continue to impose
\(B_h\widehat u_h\approx g_h\) when the prescribed boundary value differs from
those observed in training.

\paragraph{Resolution transfer.}
The resolution-transfer experiment is run on P-NS by training at \(64^2\) and
evaluating at \(128^2\). Ground-truth trajectories are generated directly by
the \(128^2\) solver, with no interpolation from coarse trajectories. The
discrete operators \(D_h\), \(P_h\), and \(R_h\) are reconstructed on the target
grid. Because cross-resolution evaluation can conflate resampling with
extrapolation to unseen frequencies and expose aliasing
\citep{sakarvadia2026false}, we use this setting only to test whether the
feasible transition remains meaningful after target-grid reconstruction; we
do not interpret it as evidence of general resolution invariance. LC
resolution transfer is kept out of the main table because it couples mesh
refinement with MAC-grid boundary projection and would add a separate
implementation factor to the stress comparison.

\paragraph{Low-data stress.}
The low-data experiment is reported in the appendix because it changes the
training set size. We retrain the methods with fixed fractions of the P-NS
training trajectories and keep the validation and test sets unchanged. This
diagnostic measures the data-efficiency effect of hard feasible transitions and
does not replace the evaluation-only stress tests reported in the main table.

\paragraph{FinalProj within the feasible-transition formulation.}
FinalProj is the closest structural control and instantiates the direct-state
realization in Eq.~\eqref{eq:posthoc_instance}, while \PCMNO{} instantiates the
projected-residual realization. Both return the predicted state to the same
solver-defined feasible set, so comparable \(\LogDiv\) or \(\LogBC\) floors
control for state-level closure. Differences in
\(\mathrm{RelL2}_{50}\) and Energy/Profile error therefore quantify the
empirical effect of transition parameterization, while
Table~\ref{tab:lc_ablation} separately examines tangent projection and
retraction under finite constraint solves.

\subsection{Hyperparameter Settings}
\label{app:hyperparameter_settings}

Table~\ref{tab:hyperparameter_settings} lists the settings used in the
paper-scale experiments. Architecture and optimizer choices are shared across
methods within each dataset, and all method-specific loss weights are fixed
before test evaluation.

For P-NS, the PINO residual weight is selected using validation rollouts from
\[
\lambda_{\mathrm{pde}}
\in
\left\{
10^{-5},
3\times10^{-5},
10^{-4},
3\times10^{-4},
10^{-3},
3\times10^{-3}
\right\}.
\]
The two strongest candidates are rerun over three seeds
\(\{0,1,2\}\). Their validation errors are
\(0.0460\pm0.0008\) for
\(\lambda_{\mathrm{pde}}=10^{-5}\) and
\(0.0465\pm0.0008\) for
\(\lambda_{\mathrm{pde}}=3\times10^{-5}\), so we select
\(\lambda_{\mathrm{pde}}=10^{-5}\).
The LC soft-loss weights are fixed before test evaluation and shared across
runs. \PCMNO{} uses only the rollout objective in
Eq.~\eqref{eq:rollout_loss} and introduces no auxiliary loss weight.

For LC, the iterative constraint solve uses \(30\) CG iterations during
training and \(120\) iterations during evaluation, with evaluation tolerance
\(10^{-6}\) and damping \(10^{-8}\). P-NS uses the closed-form spectral
Helmholtz projection implemented with FFTs.

\begin{table*}[t]
\centering
\caption{Hyperparameter settings for the paper-scale experiments. The split
sizes are numbers of trajectories. A dash indicates that the corresponding
term is absent for that constraint configuration.}
\label{tab:hyperparameter_settings}
{\scriptsize
\setlength{\tabcolsep}{5.0pt}
\renewcommand{\arraystretch}{1.05}
\begin{tabular}{@{}lcc@{}}
\toprule
Setting & P-NS & LC \\
\midrule
Spatial resolution
& \(64\times64\)
& \(64\times64\) \\
Train/validation/test trajectories
& \(500/100/200\)
& \(500/100/200\) \\
FNO layers / width / Fourier modes
& \(4/48/16\)
& \(4/48/16\) \\
Optimizer
& AdamW
& AdamW \\
Learning rate / weight decay
& \(7\times10^{-4}/10^{-4}\)
& \(7\times10^{-4}/10^{-4}\) \\
Maximum epochs
& \(100\)
& \(100\) \\
Training / evaluation batch size
& \(16/8\)
& \(8/8\) \\
Training rollout horizon
& \(H=4\)
& \(H=4\) \\
Validation / test rollout horizon
& \(20/50\)
& \(50/50\) \\
Random seeds
& \(\{0,1,2\}\)
& \(\{0,1,2\}\) \\
\midrule
DivReg/PINO divergence weight \(\lambda_{\mathrm{div}}\)
& \(0.10\)
& \(0.05\) \\
Boundary penalty weight \(\lambda_{\mathrm{bc}}\)
& --
& \(0.10\) \\
PINO residual weight \(\lambda_{\mathrm{pde}}\)
& \(10^{-5}\)
& \(10^{-2}\) \\
Constraint solve
& spectral FFT
& CG: \(30/120\) train/eval iterations \\
Evaluation solve tolerance / damping
& --
& \(10^{-6}/10^{-8}\) \\
\bottomrule
\end{tabular}}
\end{table*}

\subsection{Model and optimizer}
\label{app:model_optimizer}

All main experiments instantiate each method with the same four-layer FNO
backbone, using width \(48\), \(16\) Fourier modes per spatial dimension, and
GELU activations. The network predicts the next state directly for FNO, DivReg,
PINO, and FinalProj, and predicts a candidate update for \PCMNO{}. We optimize
all models with AdamW using learning rate \(7\times10^{-4}\), weight decay
\(10^{-4}\), and at most \(100\) epochs. The training batch size is \(16\) for
P-NS and \(8\) for LC. All methods use autoregressive windows of length \(H=4\),
and checkpoints are selected by validation rollout error. The complete training and method-specific loss settings are reported in
Table~\ref{tab:hyperparameter_settings}.

\subsection{Metric Definitions}
\label{app:metric_definitions}

This subsection defines the metrics used in
Table~\ref{tab:id_accuracy_feasibility},
Figure~\ref{fig:long_rollout_diagnostics},
Table~\ref{tab:stress_diagnostics}, and
Table~\ref{tab:lc_ablation}. For each random seed, a metric is first
aggregated over all trajectories in the corresponding test set. The paper then
reports the mean and standard deviation of the resulting per-seed values.
When a stress setting pools multiple test conditions, such as two extrapolated
Reynolds numbers, the per-seed metric is computed over the union of the
corresponding test trajectories. Condition-wise results may additionally be
reported in the appendix.

Unless otherwise stated, \(\|\cdot\|_2\) denotes the grid-weighted discrete
Euclidean norm over the relevant state, constraint, or profile degrees of
freedom. We use the fixed numerical stabilizer
\(\epsilon=10^{-6}\) throughout.

\paragraph{Rollout accuracy.}
At rollout horizon \(r\), the relative state error is
\begin{equation}
\label{eq:metric_rell2}
\mathrm{RelL2}_{r}
=
\frac{1}{N_{\mathrm{test}}}
\sum_{i=1}^{N_{\mathrm{test}}}
\frac{
    \left\|
        \widehat u^{(i)}_{h,r}
        -
        u^{(i)}_{h,r}
    \right\|_2
}{
    \left\|
        u^{(i)}_{h,r}
    \right\|_2
    +
    \epsilon
}.
\end{equation}
Here, \(u^{(i)}_{h,r}\) and \(\widehat u^{(i)}_{h,r}\) denote the reference
and predicted states, respectively, for test trajectory \(i\).
Table~\ref{tab:id_accuracy_feasibility} reports
\(r\in\{1,20,50\}\), while
Table~\ref{tab:stress_diagnostics} reports
\(\mathrm{RelL2}_{50}\). Figure~\ref{fig:long_rollout_diagnostics}
plots the same quantity as a function of the rollout horizon.

\paragraph{Constraint feasibility.}
Feasibility is evaluated using the discrete operators that define the
corresponding affine constraint set. The incompressibility metric at rollout
horizon \(r\) is
\begin{equation}
\label{eq:metric_logdiv}
\LogDiv_{r}
=
\frac{1}{N_{\mathrm{test}}}
\sum_{i=1}^{N_{\mathrm{test}}}
\log_{10}
\left(
    \left\|
        D_h\widehat u^{(i)}_{h,r}
    \right\|_2
    +
    \epsilon
\right),
\end{equation}
where \(D_h\) is the dataset-specific discrete divergence operator.

For LC, the prescribed-boundary residual is
\begin{equation}
\label{eq:metric_logbc}
\LogBC_{r}
=
\frac{1}{N_{\mathrm{test}}}
\sum_{i=1}^{N_{\mathrm{test}}}
\log_{10}
\left(
    \left\|
        B_h\widehat u^{(i)}_{h,r}
        -
        g_h^{(i)}
    \right\|_2
    +
    \epsilon
\right),
\end{equation}
where \(B_h\) extracts the constrained boundary degrees of freedom and
\(g_h^{(i)}\) contains the prescribed wall velocities for trajectory \(i\).
For P-NS, \(\LogBC_r\) is omitted because periodicity is built into the
discretization and the explicit constraint is \(C_h=D_h\).

Equations~\eqref{eq:metric_logdiv} and~\eqref{eq:metric_logbc}
average the trajectory-wise logarithmic residuals, rather than taking the
logarithm of an averaged residual. Lower and more negative values therefore
indicate stronger constraint satisfaction. The logarithmic representation is
used because FinalProj and \PCMNO{} can approach the numerical projection or
retraction floor.

\paragraph{P-NS energy error.}
For P-NS, the flow-specific diagnostic is the relative kinetic-energy error.
At rollout horizon \(r\), it is defined as
\begin{equation}
\label{eq:metric_energy_error}
\mathcal E_{\mathrm{eng},r}
=
\frac{1}{N_{\mathrm{test}}}
\sum_{i=1}^{N_{\mathrm{test}}}
\frac{
    \left|
        E\!\left(\widehat u^{(i)}_{h,r}\right)
        -
        E\!\left(u^{(i)}_{h,r}\right)
    \right|
}{
    \left|
        E\!\left(u^{(i)}_{h,r}\right)
    \right|
    +
    \epsilon
},
\end{equation}
where the discrete kinetic energy is
\begin{equation}
\label{eq:discrete_kinetic_energy}
E(v_h)
=
\frac{1}{2}
\left\|v_h\right\|_2^2.
\end{equation}

\paragraph{LC centerline profile error.}
For LC, the flow-specific diagnostic compares the two standard centerline
velocity profiles. Let
\[
\Gamma_x
=
\left\{
    (0.5,y_j)
\right\}_j
\]
denote the vertical centerline samples used for the horizontal velocity
component \(u_x\), and let
\[
\Gamma_y
=
\left\{
    (x_j,0.5)
\right\}_j
\]
denote the horizontal centerline samples used for the vertical velocity
component \(u_y\). The profile error at rollout horizon \(r\) is
\begin{equation}
\label{eq:profile_error}
\begin{aligned}
\mathcal E_{\mathrm{prof},r}
=
\frac{1}{2N_{\mathrm{test}}}
\sum_{i=1}^{N_{\mathrm{test}}}
\Bigg(
&
\frac{
    \left\|
        \left.
        \widehat u^{(i)}_{h,r,x}
        \right|_{\Gamma_x}
        -
        \left.
        u^{(i)}_{h,r,x}
        \right|_{\Gamma_x}
    \right\|_2
}{
    \left\|
        \left.
        u^{(i)}_{h,r,x}
        \right|_{\Gamma_x}
    \right\|_2
    +
    \epsilon
}
\\
&+
\frac{
    \left\|
        \left.
        \widehat u^{(i)}_{h,r,y}
        \right|_{\Gamma_y}
        -
        \left.
        u^{(i)}_{h,r,y}
        \right|_{\Gamma_y}
    \right\|_2
}{
    \left\|
        \left.
        u^{(i)}_{h,r,y}
        \right|_{\Gamma_y}
    \right\|_2
    +
    \epsilon
}
\Bigg).
\end{aligned}
\end{equation}

\begin{table*}[t]
\centering
{\scriptsize
\setlength{\tabcolsep}{3.8pt}
\renewcommand{\arraystretch}{0.98}
\begin{tabular}{@{}lclcccc@{}}
\toprule
Dataset
& Batch
& Method
& \shortstack{Step\\ms/traj. \(\downarrow\)}
& \shortstack{50-step\\ms/traj. \(\downarrow\)}
& \shortstack{Speedup vs.\\solver \(\uparrow\)}
& \shortstack{Cost vs.\\FNO \(\downarrow\)} \\
\midrule

\multirow{8}{*}{P-NS}
& 1
& Reference solver
& \(26.70 \pm 0.90\)
& \(1297.33 \pm 7.78\)
& \(1.0\times\)
& -- \\
& 1
& FNO
& \(2.62 \pm 0.29\)
& \(124.37 \pm 0.86\)
& \(10.4\times\)
& \(1.00\times\) \\
& 1
& FinalProj
& \(3.13 \pm 0.29\)
& \(147.71 \pm 0.73\)
& \(8.8\times\)
& \(1.19\times\) \\
& 1
& \PCMNO{}
& \(3.63 \pm 0.59\)
& \(170.21 \pm 1.07\)
& \(7.6\times\)
& \(1.37\times\) \\
\cmidrule(lr){2-7}
& 8
& Reference solver
& \(3.55 \pm 0.12\)
& \(174.26 \pm 2.21\)
& \(1.0\times\)
& -- \\
& 8
& FNO
& \(0.35 \pm 0.04\)
& \(15.75 \pm 0.17\)
& \(11.1\times\)
& \(1.00\times\) \\
& 8
& FinalProj
& \(0.39 \pm 0.03\)
& \(18.73 \pm 0.24\)
& \(9.3\times\)
& \(1.19\times\) \\
& 8
& \PCMNO{}
& \(0.45 \pm 0.03\)
& \(21.53 \pm 0.14\)
& \(8.1\times\)
& \(1.37\times\) \\

\midrule

\multirow{8}{*}{LC}
& 1
& Reference solver
& \(639.02 \pm 13.57\)
& \(31806.49 \pm 68.21\)
& \(1.0\times\)
& -- \\
& 1
& FNO
& \(2.69 \pm 0.49\)
& \(120.11 \pm 0.56\)
& \(264.8\times\)
& \(1.00\times\) \\
& 1
& FinalProj
& \(53.05 \pm 6.06\)
& \(2512.74 \pm 12.68\)
& \(12.7\times\)
& \(20.92\times\) \\
& 1
& \PCMNO{}
& \(97.80 \pm 0.92\)
& \(4988.19 \pm 81.61\)
& \(6.4\times\)
& \(41.53\times\) \\
\cmidrule(lr){2-7}
& 8
& Reference solver
& \(80.27 \pm 0.64\)
& \(4008.23 \pm 3.08\)
& \(1.0\times\)
& -- \\
& 8
& FNO
& \(0.33 \pm 0.06\)
& \(16.16 \pm 0.67\)
& \(248.0\times\)
& \(1.00\times\) \\
& 8
& FinalProj
& \(6.40 \pm 0.09\)
& \(316.86 \pm 0.26\)
& \(12.6\times\)
& \(19.61\times\) \\
& 8
& \PCMNO{}
& \(12.54 \pm 0.22\)
& \(618.00 \pm 4.39\)
& \(6.5\times\)
& \(38.24\times\) \\

\bottomrule
\end{tabular}}
\caption{
End-to-end computational cost on an NVIDIA A100-SXM4-40GB GPU.
One learned transition and one reference-solver call advance the state over
the same saved physical interval. P-NS uses ten pseudo-spectral RK4
micro-steps per saved step, while LC uses ten MAC micro-steps with the
data-generation constraint-solve budget. Timings include all inference-time
projection and retraction operations and exclude checkpoint loading and file
I/O. Speedup is computed against the corresponding reference solver at the
same batch size. Values are mean~\(\pm\)~standard deviation over repeated
synchronized timing runs.
}
\label{tab:end_to_end_runtime}
\end{table*}

Figure~\ref{fig:long_rollout_diagnostics} plots
\(\mathcal E_{\mathrm{eng},r}\) for P-NS and
\(\mathcal E_{\mathrm{prof},r}\) for LC across rollout horizons.
Table~\ref{tab:stress_diagnostics} reports
\(\mathcal E_{\mathrm{eng},50}\) for P-NS settings and
\(\mathcal E_{\mathrm{prof},50}\) for LC settings.
Table~\ref{tab:lc_ablation} additionally reports
\(\mathcal E_{\mathrm{prof},50}\) for the LC component variants.

\section{Additional Experiment Results}
\label{app:additional_experiment_results}

\subsection{External Structure-Preserving Baseline}
\label{app:clawno_external_baseline}

ClawNO is included in the P-NS block of
Table~\ref{tab:id_accuracy_feasibility} as an external structure-preserving
baseline \citep{liu2024clawno}. We use its FNO-based clawFNO realization,
which predicts a scalar potential \(\psi_\theta\) and constructs
\begin{equation}
\widehat u_{h,k+1}
=
\begin{bmatrix}
\partial_{y,h}\psi_\theta\\
-\partial_{x,h}\psi_\theta
\end{bmatrix},
\label{eq:clawno_pns}
\end{equation}
through fixed spectral differentiation. Hence
\(D_h\widehat u_{h,k+1}\approx0\) up to numerical differentiation error.
ClawNO therefore enforces incompressibility through the output
parameterization, whereas FinalProj projects a directly predicted state and
\PCMNO{} projects a candidate residual update before advancement and
retraction.

The matched comparison uses the same P-NS splits, \(64^2\) resolution,
Reynolds-number range, training horizon \(H=4\), and seeds \(\{0,1,2\}\).
ClawNO is trained from scratch with matched backbone capacity, while FinalProj
and \PCMNO{} reuse their validation-selected checkpoints. All methods use the
same \(50\)-step evaluator; runtime is measured at batch size one. We restrict
ClawNO to P-NS because its periodic potential representation does not directly
encode the stacked LC constraint \(C_h=[D_h;B_h]\).

Table~\ref{tab:clawno_external_baseline} supplements the main ID table with
the P-NS kinetic-energy diagnostic and inference cost.

\begin{table}[t]
\centering
{\scriptsize
\setlength{\tabcolsep}{2.4pt}
\renewcommand{\arraystretch}{0.98}
\begin{tabular}{@{}lcccc@{}}
\toprule
Method
& \(\mathrm{RelL2}_{50}\downarrow\)
& \(\LogDiv_{50}\downarrow\)
& \(\mathcal E_{\mathrm{eng},50}\downarrow\)
& \shortstack{Time/step\\(ms) \(\downarrow\)} \\
\midrule
ClawNO
& \(0.9270 \pm 0.3536\)
& \(-5.3557 \pm 0.0179\)
& \(1.1820 \pm 0.8115\)
& \(4.15 \pm 0.07\) \\
FinalProj
& \(\underline{0.0503 \pm 0.0017}\)
& \(\mathbf{-5.4981 \pm 0.0001}\)
& \(\mathbf{0.0080 \pm 0.0004}\)
& \(\mathbf{3.13 \pm 0.29}\) \\
\PCMNO{}
& \(\mathbf{0.0429 \pm 0.0005}\)
& \(\underline{-5.4978 \pm 0.0002}\)
& \(\underline{0.0086 \pm 0.0058}\)
& \(\underline{3.63 \pm 0.59}\) \\
\bottomrule
\end{tabular}}
\caption{
P-NS comparison with ClawNO. ClawNO parameterizes the velocity field through
a learned potential and fixed differentiation, while FinalProj and \PCMNO{}
retain the velocity-space backbone used in the main experiments.
}
\label{tab:clawno_external_baseline}
\end{table}

Since all three methods impose P-NS incompressibility structurally,
\(\LogDiv_{50}\) mainly verifies feasibility. The discriminating quantities
are \(\mathrm{RelL2}_{50}\), \(\mathcal E_{\mathrm{eng},50}\), and inference
cost.

\subsection{Constraint-solver diagnostics}
\label{app:projection_tolerance_runtime}

P-NS uses the closed-form spectral Helmholtz projection implemented with FFTs,
so its feasibility floor is determined by floating-point accuracy rather than
an iterative stopping rule. LC uses a MAC-grid normal-equation projection with
boundary lifting. We evaluate its numerical behavior separately because the
linear-solve budget controls a cost--feasibility tradeoff that is distinct from
neural-backbone hyperparameters.

\paragraph{Ground-truth consistency.}
\label{app:groundtruth_consistency}
Projecting held-out LC ground-truth states at rollout step \(50\) changes them
by approximately \(1.08\times10^{-7}\) in relative norm. The prescribed-wall
residual remains at \(-6.0000\) on the reported \(\log_{10}\) scale. Thus the
stored solver trajectories already lie close to the same discrete feasible set
used by the constraint layer, and the evaluation does not compare predictions
against a materially different post-processed target.

\begin{figure}[t]
\centering
\begin{minipage}[t]{0.23\textwidth}
\centering
\includegraphics[width=\linewidth]
{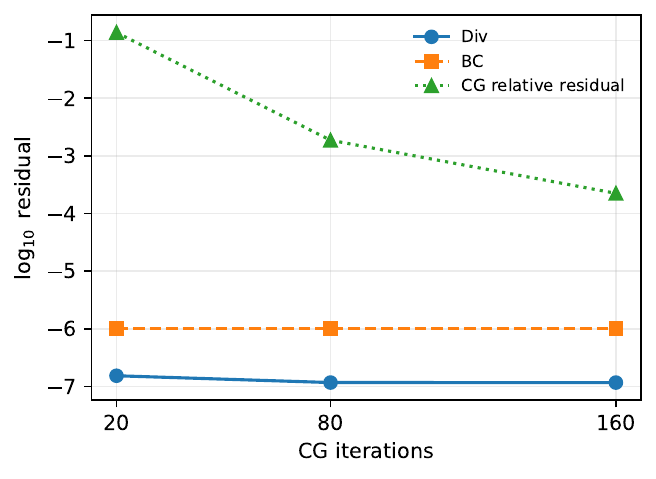}
\vspace{-0.3em}
{\scriptsize (a) Constraint and linear-solve residuals}
\end{minipage}
\hfill
\begin{minipage}[t]{0.23\textwidth}
\centering
\includegraphics[width=\linewidth]
{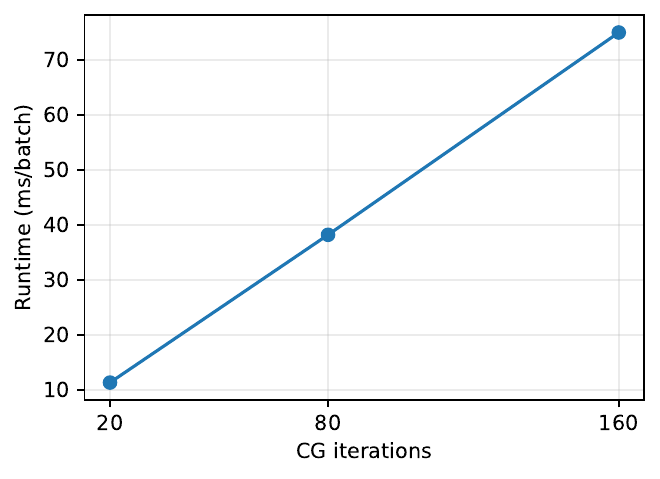}
\vspace{-0.3em}
{\scriptsize (b) Projection runtime at tolerance \(10^{-5}\)}
\end{minipage}
\caption{Numerical sensitivity of the LC projection/retraction solve on
step-\(50\) ground-truth states. At a fixed CG iteration budget, tolerances in
\(\{10^{-3},10^{-5},10^{-7}\}\) produce the same projected states, showing that
the iteration cap is the effective control in this diagnostic.}
\label{fig:projection_solver_sensitivity}
\end{figure}

Increasing the CG budget from \(20\) to \(80\) iterations reduces the interior
Div residual from \(-6.8162\) to \(-6.9341\), while \(160\) iterations yields
only a marginal further change to \(-6.9345\). BC remains at \(-6.0000\)
throughout. At tolerance \(10^{-5}\), runtime increases from \(11.34\) to
\(38.19\) and \(74.97\) ms per batch for \(20\), \(80\), and \(160\)
iterations, respectively. The paper-scale LC evaluation uses \(120\)
iterations, which lies in the saturation regime indicated by the \(80\)- and
\(160\)-iteration diagnostics. These results justify using an
intermediate-to-high solve budget rather than the highest-cost setting. The
observed saturation is consistent with Appendix~\ref{app:proof_residual_reset},
where feasibility is controlled by the achieved linear-system residual rather
than by a nominal stopping tolerance alone.

\subsection{End-to-End Computational Cost}
\label{app:end_to_end_runtime}

Table~\ref{tab:end_to_end_runtime} compares the complete learned transitions
with the corresponding reference solvers on the same GPU and over the same
saved physical interval. The measurements include all projection and
retraction operations used during inference and exclude checkpoint loading and
file I/O. DivReg and PINO share the FNO inference graph because their physical
terms are used only during training.

On P-NS, the FFT-based constraint layers introduce moderate overhead:
\PCMNO{} costs approximately \(1.37\times\) the FNO rollout while retaining a
\(7.6\times\) speedup over the reference solver at batch size one. The result
is similar at batch size eight. On LC, the iterative constraint solves dominate
the learned-model cost. FinalProj, which applies one state retraction per step,
costs approximately \(21\times\) FNO, while \PCMNO{}, which applies both
tangent projection and state retraction, costs approximately \(42\times\) FNO.
Nevertheless, the complete LC transition remains approximately
\(6.4\times\) faster than the reference MAC solver.

These measurements expose a discretization-dependent cost--feasibility
trade-off. Spectral constraints can be enforced with relatively small
overhead, whereas boundary-aware iterative solves are substantially more
expensive. The reported speedups characterize the implementations and hardware
used in this study and should not be interpreted as universal comparisons with
all optimized CFD solvers. Reducing the LC cost through warm starts,
preconditioning, or adaptive solve budgets is an important direction for
future work.

\end{document}